\pdfoutput=1
\documentclass[11pt]{article}

\PassOptionsToPackage{table}{xcolor}
\usepackage[final]{acl}
\usepackage{tablefootnote}
\usepackage{float}
\usepackage{times}
\usepackage{latexsym}
\usepackage{algorithm}
\usepackage{algorithmic}
\usepackage{booktabs}

\usepackage[T1]{fontenc}
\usepackage[utf8]{inputenc}

\usepackage{microtype}
\usepackage[most]{tcolorbox}
\usepackage[table]{xcolor}

\usepackage{inconsolata}

\usepackage{graphicx} \usepackage{multirow}
\usepackage{amssymb}
\usepackage{arydshln}

\title{Multi-Agent Simulator Drives Language Models for \\ Legal Intensive Interaction}

\author{
 \textbf{Shengbin Yue\textsuperscript{1}\protect\footnotemark[2]},
 \textbf{Ting Huang\textsuperscript{1}\protect\footnotemark[2]},
 \textbf{Zheng Jia\textsuperscript{1}\protect\footnotemark[2]},
 \textbf{Siyuan Wang\textsuperscript{2}},
  \textbf{Shujun Liu\textsuperscript{1}},
 \textbf{Yun Song\textsuperscript{3}\protect\footnotemark[1]},   \\
 \textbf{Xuanjing Huang \textsuperscript{1}},
 \textbf{Zhongyu Wei\textsuperscript{1}\protect\footnotemark[1]}
\\
 \textsuperscript{1}Fudan University, China \\
 \textsuperscript{2}University of Southern California, USA\\
 \textsuperscript{3}Northwest University of Political and Law, China
\\
\normalsize\texttt{\{sbyue23,thuang24,zjia24\}@m.fudan.edu.cn, zywei@fudan.edu.cn}
\normalsize\texttt{
}
}

\begin{document}

\maketitle
\begin{abstract}
Large Language Models (LLMs) have significantly advanced legal intelligence, but the scarcity of scenario data impedes the progress toward interactive legal scenarios.
This paper introduces a Multi-agent Legal Simulation Driver (MASER) to scalably generate synthetic data by simulating interactive legal scenarios. 
Leveraging real-legal case sources, MASER ensures the consistency of legal attributes between participants and introduces a supervisory mechanism to align participants' characters and behaviors as well as addressing distractions.
A Multi-stage Interactive Legal Evaluation (MILE) benchmark is further constructed to evaluate LLMs’ performance in dynamic legal scenarios. 
Extensive experiments confirm the effectiveness of our framework. 
% Our framework.
The detailed resources are available at \emph{\url{https://github.com/FudanDISC/MASER}}.
% \footnote{The detailed resources will be available upon acceptance.}
\end{abstract}

\renewcommand{\thefootnote}{\fnsymbol{footnote}}
\footnotetext[2]{These authors contributed equally to this work.}
\footnotetext[1]{Corresponding author}
\renewcommand{\thefootnote}{\arabic{footnote}}

\section{Introduction} 
The emergence of Legal Artificial Intelligence (Legal AI)~\cite{atkinson2020explanation,ge2021learning} induces significant transformations within the legal field.
Especially, recent advancements in Large Language Models (LLMs)~\cite{achiam2023gpt, yao2024lawyer,yue2024lawllm}, through continuous training on legal knowledge and instructions, have shown remarkable capabilities in various legal tasks, such as
legal information extraction~\cite{shen2024empowering},
judgment prediction~\cite{he2024simucourt} and instrument drafting~\cite{yue2023disc}. 
However, previous endeavors ~\cite{fei2023lawbench} predominately concentrated on non-intensive interactive tasks, involving prediction with directly specified elements, thereby restricting the wider real-world application of Legal AI.

\begin{figure}[t]
    \centering
\includegraphics[width=\linewidth]{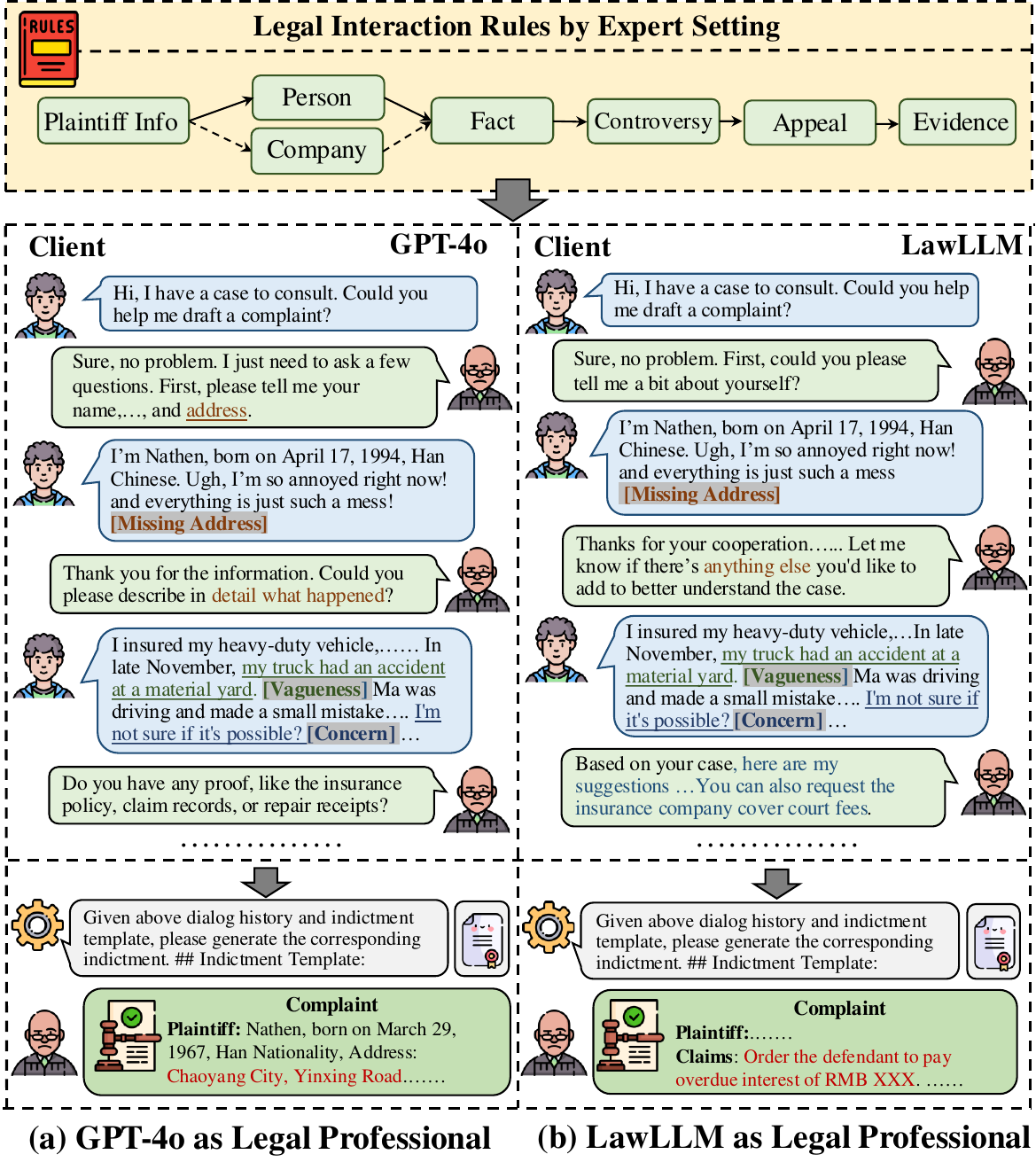}
    \caption{ Examples of general LLM (\textit{i.e.}, GPT-4o) and legal LLM (\textit{i.e.}, LawLLM) as legal professionals in drafting legal documents. LLMs struggle to maintain flexible interaction patterns under legal agendas.}
    \label{fig:intro}
\end{figure}

The real-world judicial service scenario is highly complex, presenting two major challenges: (1) High professionalism. The rationality of content and the legality of procedures are crucial for legal information services, requiring practitioners to follow a procedural agenda while offering services. (2) Intensive interactivity. Since users often lack basic legal knowledge, legal consultations typically require multiple rounds of interaction, with practitioners guiding users step by step through the process. 
As shown in Figure \ref{fig:intro}, drafting a legal complaint is a complex task that demands fulfilling all the listed requirements.
However, when the user exhibits incomplete information, vagueness or concerns, both GPT-4o \cite{achiam2023gpt} and LawLLM \cite{yue2024lawllm} struggle with adaptive handling while simultaneously following agendas. 
This results in misunderstanding of the user's demands and omission of necessary facts, ultimately inadequately drafting complaints.
Current LLMs face challenges in this dynamic legal scenario.
A fundamental fact lies in the proprietary nature and scarcity of such data in real legal scenarios, which creates substantial barriers to equipping LLMs with such advanced capabilities.

Therefore, we explore a role-driven scenario simulation framework to construct high-quality data in scale for enhancing legal LLMs. 
Compared existing role-driven approaches for general contexts~\cite{shao2023character,yu2024beyond,yu2024large}, two key aspects require consideration to better model real-world legal scenarios.
(1) \emph{Character Authenticity}. The legal simulation involves various roles, including legal characters with expert knowledge, solutions and agendas, as well as non-legal characters, who may exhibit diverse behavior styles and backgrounds. 
Due to LLM's susceptibility to legal illusions, legal configurations of roles (non-legal roles' needs and legal roles' solutions) require consistency with real-word scenarios, ensuring the authenticity and reliability of legal services.
(2) \emph{Character-behavior Consistency}. 
The response behavior of each role should consistently align with their characteristic throughout the interaction. For example, non-legal roles should respond according to their profiles while legal roles should always follow legal knowledge and agendas.
Additionally, non-legal roles may present distracting behaviors (forgetfulness or misunderstanding), leading to vague responses, while legal roles must respond and direct professionally.

We present a framework to address these aspects, forming the \textbf{Multi-agent Legal Simulation Driver} (\textbf{MASER}), where the legal agent (Lawyer) guides and questions the non-legal agent (Client) to complete specific legal tasks in accordance with legal agendas.
MASER incorporates two strategies:
(1) \textit{Role Agent Presetting}: 
we map multi-level traits based on a sociological personality theory to ensure diversity in non-legal roles.
To ensure legal correspondence, we extracted different elements from judicial documents (which establish a complete logical chain of legal events) to develop the legal traits of the roles. This enables the legal role to have prior knowledge of addressing user demands.
(2) \textit{Multi-Agent Legal Simulation}: 
Beyond the legal agent (Lawyer) and non-legal agent (Client), we introduce a supervisor for character-behavior and distractor alignment at the sentence level, assisting participants in revising their behaviors in each turn.
MASER leverages this multi-agent simulator as a synthetic data engine, to drive arbitrary offline LLMs for dynamic legal interaction.

Moreover, we propose a \textbf{Multi-Stage Interactive Legal Evaluation} (\textbf{MILE}) benchmark, to assess LLM variants' ability to navigate interactive legal tasks.
This benchmark is derived from a meticulous collection of high-quality Chinese legal cases, providing real-world and consistent legal configurations. 
Leveraging GPT-4o to simulate non-legal characters (client), we conduct a thorough evaluation of the performance of LLMs-driven Lawyer within this dynamic and realistic legal interaction environment. 
A supervisor oversees the client to ensure the consistency of its behavior. The MILE benchmark evaluates the Lawyer along two stages of intensive interaction and goal achievement.

Extensive experiments demonstrate that MASER significantly enhances the performance of arbitrary LLMs on interactive legal tasks. In our framework, trained models surpassed proprietary advanced LLM, \textit{e.g.}, GPT-4o~\cite{achiam2023gpt}, and specialized legal LLM, \textit{e.g.}, LawLLM~\cite{yue2024lawllm}.
We envision our framework as a general paradigm that extends more complex and private verticals, bridging the gap between intensive interaction and achieving special goals.

\begin{figure*}[t]
    \centering
\includegraphics[width=\linewidth]{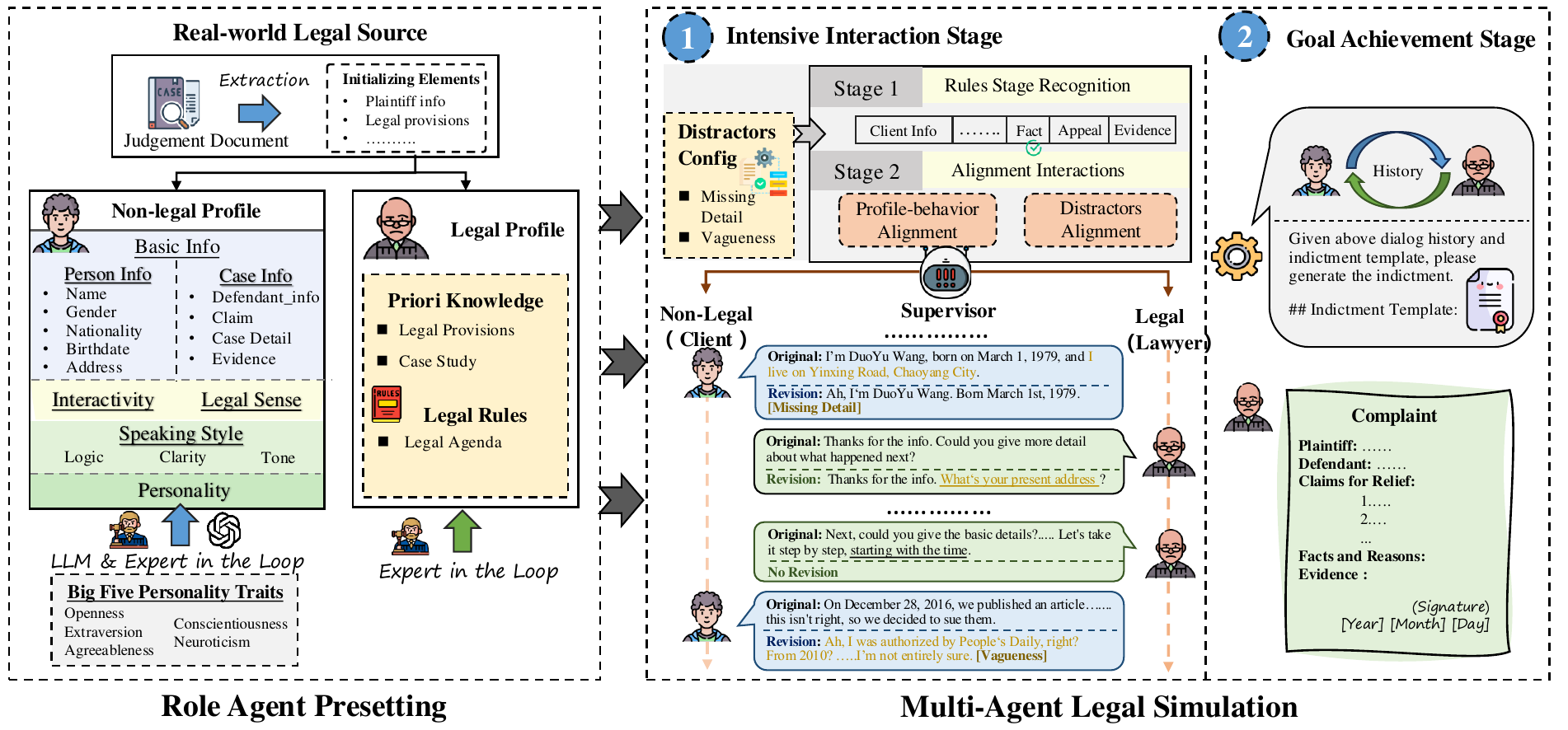}
    \caption{Overview of Multi-agent Legal Simulation Driver (MASER), which consists of role agent presetting and multi-agent legal simulation. Leveraging the MASER, synthesized sentence-level data can drive arbitrary LLMs for legal intensive interaction.}
    \label{fig:model}
\end{figure*}

\section{Multi-agent Legal Simulation Driver} 
As shown in Figure \ref{fig:model}, MASER achieves legal objectives (\textit{i.e.}, complaint drafting) through intensive interactions, which consist of three agents: a client, a lawyer, and a supervisor.
Each character assumes specific roles and responsibilities within the framework. 
The MASER consists of \textit{Role Agent Presetting} (\textsection \ref{sec:Presetting})  and \textit{Multi-Agent Legal Simulation} (\textsection \ref{sec: MALS}),  ultimately producing synthetic scene data to train arbitrary LLMs (\textsection \ref{sub:Finetuning}).

\subsection{Role Agent Presetting.}
\label{sec:Presetting}
Due to limited expertise, vanilla LLM-driven simulators tend to produce substantial legal illusions, compromising reliability of synthetic data. 
Hence, the authenticity setting of role agent 
presents two challenges: the client's diversity and legal consistency among roles. To this end, we develop the individual's features across different dimensions based on the Big-5 Personality Traits. Additionally, we configure various legal elements in real legal events with the same logic chain to different agents. 

\paragraph{Setup with Real-world Legal Source.} 
To establish legal correspondence, the intuition is to assign different legal elements of the same legal event to various roles. Chinese judgment documents encompass all the essential components of the entire proceedings of the case, \textit{e.g.}, parties' personal details, legal disputes, facts, evidence, rulings, and applicable laws, which align perfectly with our requirements. 
We delicately capture these legal elements from documents as legal initialization attributes for lawyers and clients. Specifically, we collect 4,532 civil cases from pre-2021 judgment documents, covering more than 230 civil case types. Then, we prompt GPT-4o to extract six initialization elements from Chinese judgment documents: plaintiff information, defendant information, claim, case details, evidence, case study, and legal provisions. 
Details are provided in the Appendix \ref{sec:appendix: dataset_jd}.

\paragraph{Setup with Big-5 Personality Traits.}
Behavioral diversity facilitates the model's ability to comprehend and represent the complexity of characters, further improving simulation reliability and generalization. Guided by social personality theory \cite{sun2024identity}, Big-5 personality traits are mapped to two dimensions of behavioral attributes: speaking style and interactivity, where speaking style consists of logic, clarity and tone.
In our implementation,
three levels (high, medium, or low) are assigned to each of the big-5 personality traits,
then we prompt GPT-4 to generate personality descriptions and further develop speaking style and interactivity. 
Details are shown in the Appendix \ref{sec:appendix: dataset_pt}.

\paragraph{Roles Profile Configuration.}
As shown in Figure \ref{fig:model}, these categories of elements from the above steps are assigned to the corresponding agents.
Specifically, 
(1) \textit{Client} is equipped with personal and case information from real-world legal sources, along with personality, speech styles, and interactivity from Big-5 personality traits.
Additionally, five levels of legal sense are manually designed by legal experts, aiming to model the level of legal knowledge.
(2) \textit{Lawyer} is configured with case analysis and applicable laws from the same legal source, which is the prior knowledge of addressing the client's demands. 
In addition, legal agendas are manually designed by legal experts.
(3) \textit{Supervisor} owns all the information and mounts it on demand, \textit{e.g.}, case information and personality features are configured when supervising Client's behavior.

\subsection{Multi-Agent Legal Simulation.}
\label{sec: MALS}
As shown in Figure \ref{fig:model},
the lawyer guides the client to collect legal needs, ultimately completing the user's legal task goal (\textit{i.e.}, complaint drafting). The Supervisor oversees interactions between Client and Lawyer at the sentence level, guaranteeing profiles-behavior alignment and distractor consistency.
The simulation consists of intensive interaction and goal achievement stages. We leverage powerful LLM (\textit{e.g.}, GPT-4o) to power each agent, enabling them to embody their roles authentically.

\paragraph{Client Agent Behavior.}
The Client aims to exhibit a set of realistic behavior patterns to enhance the authenticity of the legal simulation: 
1) \textit{Cooperation}. The agent should respond to the lawyer’s inquiries, and have subjective biases in describing their condition according to setting legal sense; 
2) \textit{Communication}. The agent possesses an individualized speaking style, combining logic, clarity, and emotional response; 
3) \textit{Curiosity}. The agent should express concerns based on their level of understanding and interactivity, seeking clear explanations from the lawyer to address their doubts; 4) \textit{Distraction behavior}. The agent's response exhibits two types of distractors, missing some important details or vagueness, requiring further confirmation to clarify. 
The prompt is shown in Figure \ref{fig:p2}.

\paragraph{Lawyer Agent Behavior.}
The Lawyer agent aims to emulate a skilled and patient legal service provider in real-world practice: 
1) \textit{Agenda compliance}.
Throughout the interaction, the agent must adhere to the given legal rule and agenda to guide Client, ensuring the acquisition of all elements necessary to achieve the legal objectives.
2) \textit{Flexible reaction}. The agent uses the configured prior knowledge of the case to address the user's confusion expertly, and when the user exhibits distracting behavior, further inquiry is required to ascertain the facts.
The Lawyer's prompt is shown in Figure \ref{fig:p4}.

\paragraph{Supervisor Agent Behavior.}
The Supervisor agent oversees multi-turn conversations between the Client and the Lawyer at the sentence level, improving their consistency and interactivity: 1) \textit{Profile-behavior alignment}.
The agent assesses the consistency between each participant's profile and their speech, in conjunction with providing advice based on diverse styles of the Client and legal agendas of the Lawyer.
2) \textit{Distractors alignment}.
Relying solely on profiles to prompt the Client's distracted behavior and lawyers' flexible reaction is difficult. Thus, the agents use preset distractor configurations to guide participants' interactions under distraction, \textit{e.g.}, the Client missing details and the Lawyer queries again.
The Supervisor first determines the current agenda stage by identifying conversations, and subsequently implementing appropriate supervisory measures according to the stage.
The supervisor provides feedback and modification suggestions for the Client or Lawyer to adjust their responses with natural language.
The prompts are shown in Figure \ref{fig:p5}, \ref{fig:p6} and \ref{fig:p10}.

\begin{algorithm}[tb]
\caption{ Simulation Process}
\label{algo:Supervision Process}
\textbf{Require}: Lawyer 
$\mathcal{L}$, Supervisor $\mathcal{S}$, Client $\mathcal{C}$, Max-turn $m$, Dialogue history $h$,  Client's response $r^{c}$,  Lawyer's response $r^{l}$, Supervisor's response $r^{s}$, Complaint template $y_{temp}$\\
\textbf{Input:} Client's profile $P_{c}$, Lawyer's profile $P_{l}$\\  
\textbf{Output:} History $h_{e}$, complaint $y$ 

\begin{algorithmic}[1]
\FOR{each $t$ in ${0,1,...,m}$}
    \STATE $\mathcal{C}$ generates $r_{before}^{c}$
    \STATE $\mathcal{S}$ generates $r^{s}$ given $P_{c}$,$h_{t-1}$,$r_{before}^{c}$
    \IF{$r^{s} = \text{"correct"}$}
    \STATE $\mathcal{C}$ memorizes $r_{before}^{c}$ in $h_{t-1}$
    \ELSE
    \STATE $\mathcal{C}$ generates $r_{after}^{c}$ given $r^{s}$, $r_{before}^{c}$
    \STATE $\mathcal{C}$ memorizes $r_{after}^{c}$ in $h_{t-1}$
    \ENDIF
    
    \STATE $\mathcal{L}$ generates $r^{l}_{before}$ given $h_{t-1}$
    \STATE $\mathcal{S}$ generates $r^{s}$ given $P_{c}$, $P_{l}$, $h$, $r^{l}_{before}$
    \IF{$r^{s} = \text{"correct"}$}
    \STATE $\mathcal{L}$ memorizes $r^{l}_{before}$ in $h_{t-1}$
    \ELSE
    \STATE $\mathcal{L}$ generates $r^{l}_{after}$ given $r^{s}$, $r^{l}_{before}$
    \STATE $\mathcal{L}$ memorizes $r^{l}_{after}$ in $h_{t-1}$
    \ENDIF  
    \IF{$\text{"Inquiry ends"}$ in $r^{l}$}
    \STATE break
    \ENDIF
\ENDFOR
\STATE $\mathcal{L}$ predicts $y$ given $h_{m}$, $y_{temp}$
\STATE $h_{e}=\mathcal{L}$ memorizes $y$ in $h_{m}$
\end{algorithmic}
\end{algorithm}
\paragraph{Legal Simulation Flow.}
As shown in Figure \ref{fig:model},
our framework simulates a realistic complaint drafting process through a structured multi-turn interaction flow. 
The Client initiates the conversation based on the configured real-case cause. 
The lawyer agent acts as the dialogue facilitator, engaging in interactions with the Client directed by agendas.
Across interactions, the Supervisor interacts only with the current speaker in each turn.
As shown in Algorithm \ref{algo:Supervision Process}, 
in $t$-th round, after Client $\mathcal{C}$ (or Lawyer $\mathcal{L}$) generates a response $r^{c}_{before}$, the Supervisor $\mathcal{S}$ provides suggestions $r^{s}$ based on the previous dialogue $h_{t-1}$ and Client's profile $P_{c}$. If $r^{s}$ is deemed "correct," $r^{c}_{before}$ is added to memory; Otherwise, the Client (or the Lawyer) revises their response $r^{c}_{after}$ according to the Supervisor's suggestions $r^{s}$ before adding it to memory. 
The interaction ends when $\mathcal{L}$ reaches the inquiry end or the predefined maximum turn $m$. Ultimately, $\mathcal{L}$ generates the complaint $y$ based on the complete history $h_{m}$ and template $y_{temp}$, simultaneously updating it into $h_{m}$ to form scenario data $h_{e}$.
Figure \ref{fig:p17} shows an example simulation flow.
\label{sub:Simulation}

\subsection{Training}
\label{sub:Finetuning}
\paragraph{Synthetic Data Generation.}
MASER is utilized to construct a high-quality
synthetic legal scene dataset, \textbf{SynthLaw}, consisting of 4,532 samples. 
Note two keys in the synthetic process:
1) real-legal source configurations and supervision mechanisms in each interaction ensure that the generated data is aligned at the sentence level, closely approximating real-world scenarios.
2) the diverse client behavioral styles and legal demands ensure the data generalization.
This approach greatly remedies the dilemma of scene data construction under legal resources.
In addition, we collect 4k high-quality multi-round legal counseling data to enhance the legal routine Q\&A capability. 

\paragraph{Supervised Finetuning.}
We initialize a general LLM and train it on SynthLaw dataset $D_{s}$. For each example $\left(X_{c}^{1}, X_{l}^{1}, \ldots, X_{c}^{T}, X_{l}^{T}\right)\subset D_{s}$,
where $c$ and $l$ denote Client and Lawyer, a standard conditional language modeling objective, maximizing likelihood:
$$
\mathcal{L}=-\sum_{i=1}^{L} \log p\left(x_{i} \mid X_{c,<i}, X_{l,<i}\right), x_{i} \in X_{r}
$$
\noindent where $L$ is the token length of sequence $X$, $x_{i}$ is the current predicted the Lawyer's response tokens, $X_{c,<i}$ and $X_{l,<i}$ are the Client's response and Lawyer's response tokens before $x_{i}$.

\section{MILE Benchmark}
\begin{table*}[]
\centering
\resizebox{0.8\textwidth}{!}{
\begin{tabular}{ccccccccc}
\cline{1-9}
\multicolumn{1}{c|}{\multirow{2}{*}{Model}} & \multicolumn{5}{c|}{Local}                      & \multicolumn{2}{c|}{Global} & \multirow{2}{*}{AVE} \\
\multicolumn{1}{c|}{}                       & CLI   & DEF   & F \& R & CLA   & \multicolumn{1}{c|}{EVID} & STA               & \multicolumn{1}{c|}{PROF}              &                          \\ \hline
\rowcolor{gray!25}
\multicolumn{9}{c}{Multilingual LLMs}                                                                                             \\
\multicolumn{1}{l|}{GPT-4o}                 & \textbf{94.26} & \textbf{98.25} & 40.39  & \underline{66.00} & \underline{52.57}                    & \underline{84.63}             &  \multicolumn{1}{c|}{\textbf{63.95}}            & \underline{71.44}                    \\
\multicolumn{1}{l|}{GPT-3.5-turbo}                &      85.23 &  92.31  & 36.33 & 55.31 &  45.06   &          39.44    &       \multicolumn{1}{c|}{48.23}      &   51.77               \\
\multicolumn{1}{l|}{Gemini-1.5-pro}         & 55.95 & 58.45 & 31.60  & 41.57 & 20.09                     & 44.81             &  \multicolumn{1}{c|}{ 38.34}             & 39.96                    \\\hdashline
\multicolumn{1}{l|}{Baichuan2-chat\space$_{\mathrm{13B} }$}  & 78.64 & 74.69 & 40.35  & 48.33 & 29.22      & 62.02             &        \multicolumn{1}{c|}{43.82}        & 53.87          \\
\multicolumn{1}{l|}{LLaMa-3.1-instruct\space$_{\mathrm{8B} }$}  & 85.33 & 93.11 & \underline{41.93}  & 57.81 & 27.72                     & 56.64             &        \multicolumn{1}{c|}{46.47}        & 52.14                    \\
\multicolumn{1}{l|}{Baichuan2-chat\space$_{\mathrm{7B} }$}  & 79.37 & 78.44 & 34.21  & 47.94 & 29.39                     & 55.73             & \multicolumn{1}{c|}{44.46}             & 46.20                    \\
\multicolumn{1}{l|}{InternLM2.5-chat\space$_{\mathrm{7B} }$}            & 62.77 & 64.28 & 35.74  & 44.62 & 31.03                     & 50.97             & \multicolumn{1}{c|}{44.21}             & 43.51                    \\
\multicolumn{1}{l|}{Mistral-instruct-v0.3\space$_{\mathrm{7B} }$}                & 14.48 & 12.82 & 21.36  & 30.00    & 27.60                     & 23.85             & \multicolumn{1}{c|}{23.12}             & 26.56                    \\
 \hline
\rowcolor{gray!25}
\multicolumn{9}{c}{Legal LLMs}                                                                                                                                              \\
\multicolumn{1}{l|}
{LawLLM\space$_{\mathrm{13B} }$}                 & 49.91 & 38.49 & 30.49  & 32.61 & 27.24                     & 50.13             & \multicolumn{1}{c|}{27.58}             & 30.09                    \\
\multicolumn{1}{l|}{Interrogatory\space$_{\mathrm{7B} }$}    & 12.97 & 18.12 & 29.64  & 32.09 & 28.41                     & 40.45             & \multicolumn{1}{c|}{28.05}             & 30.07                    \\
\multicolumn{1}{l|}
{Fuzi.mingcha\space$_{\mathrm{6B} }$}           & 53.34 & 5.91  & 19.86  & 25.82 & 27.46                     & 24.29             & \multicolumn{1}{c|}{23.64}             & 24.73                    \\

 \hline
 \multicolumn{1}{l|}{Qwen2.5-instruct\space$_{\mathrm{7B} }$  }    & 86.90 & 92.05 & 37.81  & 55.66 & 34.23                     & 41.50             &  \multicolumn{1}{c|}{56.09}            & 57.75                    \\ \hdashline
\multicolumn{1}{l|}
{SynthLaw\space$_{\mathrm{7B} }$}                                        & \underline{90.20} & \underline{96.15} & \textbf{54.20}  & \textbf{70.00} & \textbf{54.60} & \textbf{91.74}             & \multicolumn{1}{c|}{\underline{59.05}}             & \textbf{73.71}                    \\ \hline
\end{tabular}}
\caption{Comparative results among LLMs on goal evaluation, where CLI, DEF, F\&R, CLA, and EVID denote client information, defendant information, fact reason, claims, and evidence, respectively. STA and PROF denote Standardability and Professionality, respectively. }
\label{tab:final_goal}
\end{table*}
Unlike existing legal benchmarks \cite{fei2023lawbench,yue2023disc} that employ static assessment, Multi-Stage Interactive Legal Evaluation (MILE) introduces an approach for assessing the model's ability to complete designated legal tasks in a dynamic environment. 
This benchmark offers the key advantage that it better aligns with real-world conditions and thus more reliably reflects the model's performance. 
Leveraging powerful LLM to simulate the non-legal characters (\textit{i.e.}, Client), MILE thoroughly evaluates the performance of LLMs-driven lawyer within this dynamic legal interaction environment. 
MILE is divided into two phases: 
interaction evaluation and goal evaluation. 

\begin{figure}[t]
    \centering
\includegraphics[width=0.41\textwidth]{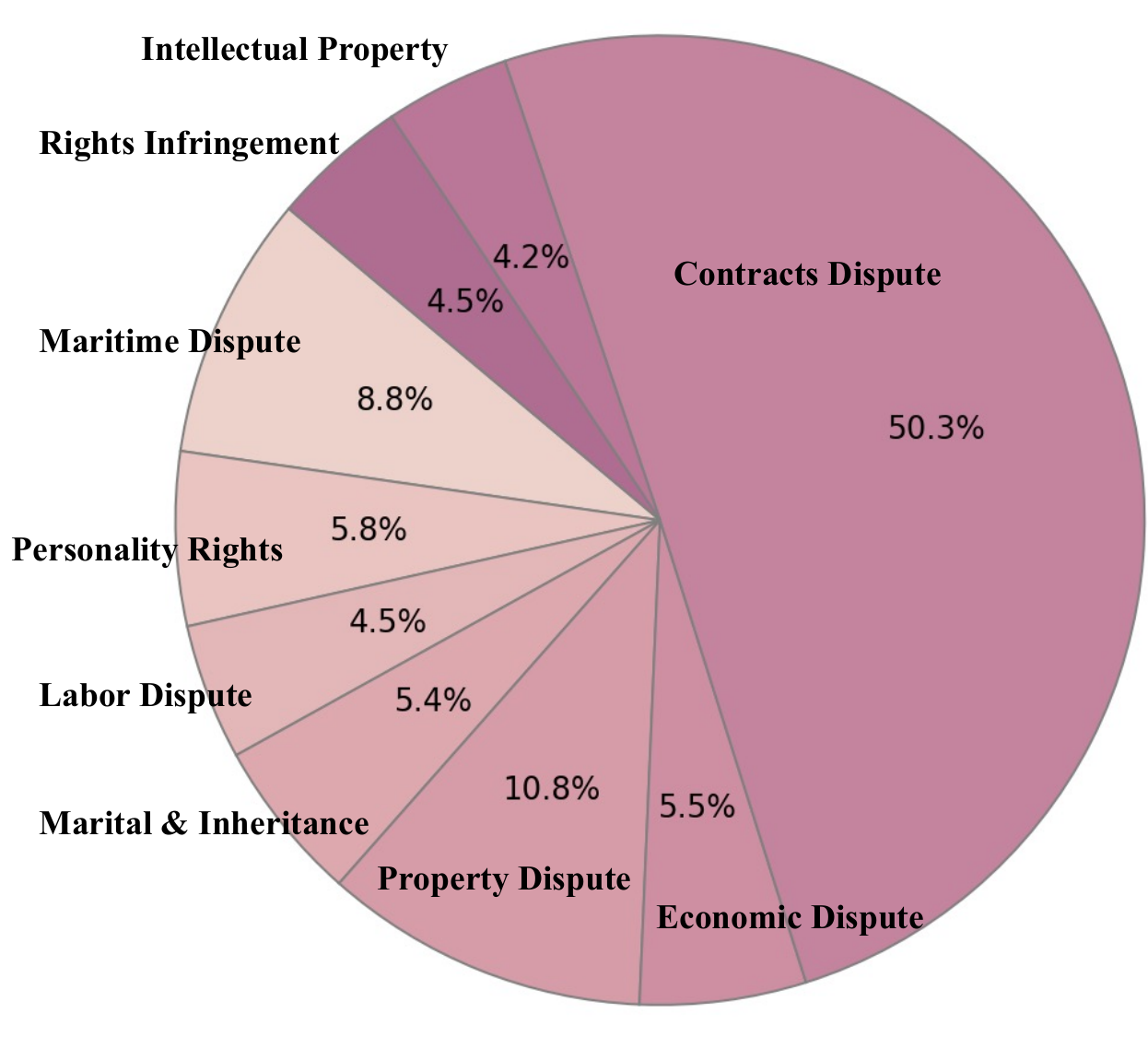}
    \caption{Distribution of legal attributes for our MILE benchmark, including 9 primary attributes.}
    \label{fig:dist}
    % \vspace{-3mm}
\end{figure}
\paragraph{Dataset Construction.}
We collect civil judgment documents from the China Judgments Online of the year 2024, and further performed privacy removal and data cleaning. 
The legal elements and behavioral styles processed by GPT-4o serve as the client's profiles. 
In total, the MILE benchmark sets out 693 distinct complaint drafting scenarios, where the complaint documents are generated from judgment documents through the heuristic method. 
Figure \ref{fig:dist} illustrates the legal attributes of our MILE.
\paragraph{Interaction Evaluation.}
This phase aims to evaluate the model's interactive performance as a lawyer, focusing on the following three aspects: 1) \emph{Interactivity}. The model should actively engage in the dialogue, answering and asking questions to advance the discussion while clarifying any vague responses.
2) \emph{Professionality}. The model should use precise legal terms, cite laws and precedents, and offer professional strategies. 
3) \emph{Logicality}. The model should maintain logical dialogue. 
Powerful judge model (\textit{i.e.}, GPT-4o) measures scores on a scale of 1 to 10.
Note that we use two turns as a window for fine-grained evaluation rather than directly evaluating the entire conversation.

\paragraph{Goal Evaluation.}
This phase evaluates the performance of the final task (\textit{i.e.,} complaint quality) from two perspectives: 1) \emph{Local} evaluates the accuracy of each part of the generated complaint, including client information, defendant information, facts, reason, claims, and evidence.
2) \emph{Global} assesses the overall standardability (whether the document follows a given template) and professionalism (whether correct legal language is used) of the complaint. 
The accuracy of client and defendant information is measured through matching, while other elements are measured by GPT-4o. For each complaint, a ground truth is provided to reduce potential biases during the assessment.
Details are provided in the Appendix \ref{sec:appendix: goal_evaluation}.

\section{Experimental Setup}

\paragraph{Implementation Detail.} 
We use Qwen2.5-instruct-7B \cite{yang2024qwen2} as our initial model. Due to page limitations, details of the training and evaluation processes are provided in Appendix \ref{sec:appendix: setting_details}.

\paragraph{Baselines.}
We compare our model with a wide range of baseline methods in two categories. 
(1) \textit{General multilingual LLMs}:
Qwen2.5-instruct 7B \cite{yang2024qwen2}, Baichuan2-chat 7B/13B \cite{yang2023baichuan}, InternLM2.5-chat 7B \cite{cai2024InternLM2}, LLaMa-3.1-instruct 8B  \cite{dubey2024llama}, Mistral-instruct-v0.3 7B \cite{jiang2023mistral}, GPT-4o  \cite{achiam2023gpt}, GPT-3.5-turbo \cite{achiam2023gpt}, Gemini-1.5-pro  \cite{reid2024gemini}.
(2) \textit{Legal-domain LLMs}: 
LawLLM \cite{yue2023disc}, Interrogatory\footnote{\url{https://github.com/zhihaiLLM/Interrogatory}}, Fuzi.mingcha\footnote{\url{https://github.com/irlab-sdu/fuzi.mingcha}}.

\section{Experiment Results}

\subsection{Main Results}
\paragraph{Comparison on goal evaluation.}
We conduct the goal evaluation from global and local perspectives, with the former perspective evaluating the generated complaint as a whole by scoring its format and professionalism, and the latter focusing on specific parts within the generated complaint. 
From the table \ref{tab:final_goal}, we observe that: 1) \textit{Comparison with the Baseline}. our SynthLaw surpasses baseline (\textit{i.e.}, Qwen2.5-instruct-7B) by a large margin on all metrics, STA in particular, which demonstrates that our framework has significantly improved the model's ability of the baseline model to achieve all legal objectives, including following the specific legal format.
2) \textit{Comparison with multilingual LLMs}. Our model surpasses multilingual models of the same size on all metrics. Even compared to closed-source LLMs trained on private data, our model outperforms them in most metrics. Particularly, our model exceeds GPT-4o in terms of overall average scores.
3) \textit{Comparison with legal LLMs.} 
Although these domain-specific LLMs perform better
than general LLMs on legal tasks \cite{yue2023disc,yue2024lawllm}, they lack interactive skills to identify elements such as relevant facts and evidence, resulting in low scores in local evaluation. 

\begin{table}[]
\resizebox{\columnwidth}{!}{%
\begin{tabular}{ccccc}
\hline
\multicolumn{1}{c|}{Model}                  & INT   & PROF  & \multicolumn{1}{c|}{LOGI}  & AVE   \\ \hline
\rowcolor{gray!25}
\multicolumn{5}{c}{Multilingual LLMs}                                                            \\ \hline
\multicolumn{1}{l|}{GPT-4o}                 & \cellcolor[HTML]{e4f8e4}82.21 & \cellcolor[HTML]{a5daa6}\underline{74.70} & \multicolumn{1}{c|}{\cellcolor[HTML]{a5daa6}\underline{79.69}} & \cellcolor[HTML]{e4f8e4}78.86 \\
\multicolumn{1}{l|}{GPT-3.5-turbo}          & 78.01 & 71.62 & \multicolumn{1}{c|}{76.83} & 75.49 \\
\multicolumn{1}{l|}{Gemini-1.5-pro}        & \cellcolor[HTML]{5eb95f}\textbf{83.46} & \cellcolor[HTML]{5eb95f}\textbf{75.05} & \multicolumn{1}{c|}{\cellcolor[HTML]{5eb95f}\textbf{79.82}} & \cellcolor[HTML]{5eb95f}\textbf{79.45} \\ \hdashline
\multicolumn{1}{l|}{Baichuan2-chat\space$_{\mathrm{13B} }$}  & 72.58 & 65.33 & \multicolumn{1}{c|}{70.37} & 69.42 \\
\multicolumn{1}{l|}{LLaMa-3.1-inst.\space$_{\mathrm{8B} }$}  & 79.29 & 71.94 & \multicolumn{1}{c|}{76.98} & 76.07 \\
\multicolumn{1}{l|}{Baichuan2-chat\space$_{\mathrm{7B} }$} & 71.62 & 64.21 & \multicolumn{1}{l|}{69.62} & 68.43 \\
\multicolumn{1}{l|}{InternLM2.5-chat\space$_{\mathrm{7B} }$}              & 64.35 & 60.18 & \multicolumn{1}{c|}{63.81} & 62.78 \\
\multicolumn{1}{l|}{Mistral-inst.-v0.3\space$_{\mathrm{7B} }$}              & 21.64 & 24.84 & \multicolumn{1}{c|}{23.82} & 23.43 \\\hline 
\rowcolor{gray!25}
\multicolumn{5}{c}{Legal LLMs}                                                                  \\ \hline
\multicolumn{1}{l|}{LawLLM\space$_{\mathrm{13B} }$}         & 57.25 & 53.17 & \multicolumn{1}{c|}{56.42} & 55.61 \\
\multicolumn{1}{l|}{Interrogatory\space$_{\mathrm{7B} }$}    & 52.95 & 49.56 & \multicolumn{1}{c|}{52.82} & 51.78 \\
\multicolumn{1}{l|}{Fuzi.mingcha\space$_{\mathrm{6B} }$}           & 51.52 & 46.53 & \multicolumn{1}{c|}{50.34} & 49.47 \\ \hline
\multicolumn{1}{l|}{Qwen2.5inst.\space$_{\mathrm{7B} }$}   & 72.90 & 68.46 & \multicolumn{1}{c|}{72.73} & 71.36 \\  \hdashline
\multicolumn{1}{l|}{SynthLaw\space$_{\mathrm{7B} }$}              &   \cellcolor[HTML]{a5daa6}\underline{83.23}    &  \cellcolor[HTML]{e4f8e4} \textbf{74.48}    & \multicolumn{1}{c|}{\cellcolor[HTML]{e4f8e4}\textbf{79.22}}      &   \cellcolor[HTML]{a5daa6}\underline{78.97}    \\ \hline
\end{tabular}%
}
\caption{Comparative results among LLMs on interaction evaluation, where INT, PROF, LOGI denote Interactivity, Professionality, and Logicality. Darker (best) to lighter green marks the best of the top three results.}
\label{tab:exp_ie}
\end{table}

\paragraph{Comparison on interaction evaluation.} 
We take two rounds as a window to assess the interaction process turn-by-turn and ultimately calculate the average score for each metric. 
As shown in Table \ref{tab:exp_ie}, our SynthLaw improved by 14.17\%, 8.79\% and 8.92\% in average scores of INT, PROF and LOGI, compared to the vanilla LLM.
Note that the performance of legal LLMs is weaker than that of general-purpose LLMs, further demonstrating their limitations in the interactive capabilities.
While our model significantly outperforms current legal LLMs, its performance is slightly lacking when compared to the proprietary Gemini-1.5-pro which has undergone extensive alignment and fine-tuning.
Given the size of our model and the volume of the training data, this limitation appears to be reasonable.
Nevertheless, their performance on the final tasks is inferior to ours, which can further prove the effectiveness of our model. 
The experiments demonstrate that our approach can enhance the dense interaction capabilities of existing offline models, bridging the gap between intensive interaction and achieving legal goals.

\begin{figure}[t]
    \centering
\includegraphics[width=0.5\textwidth]{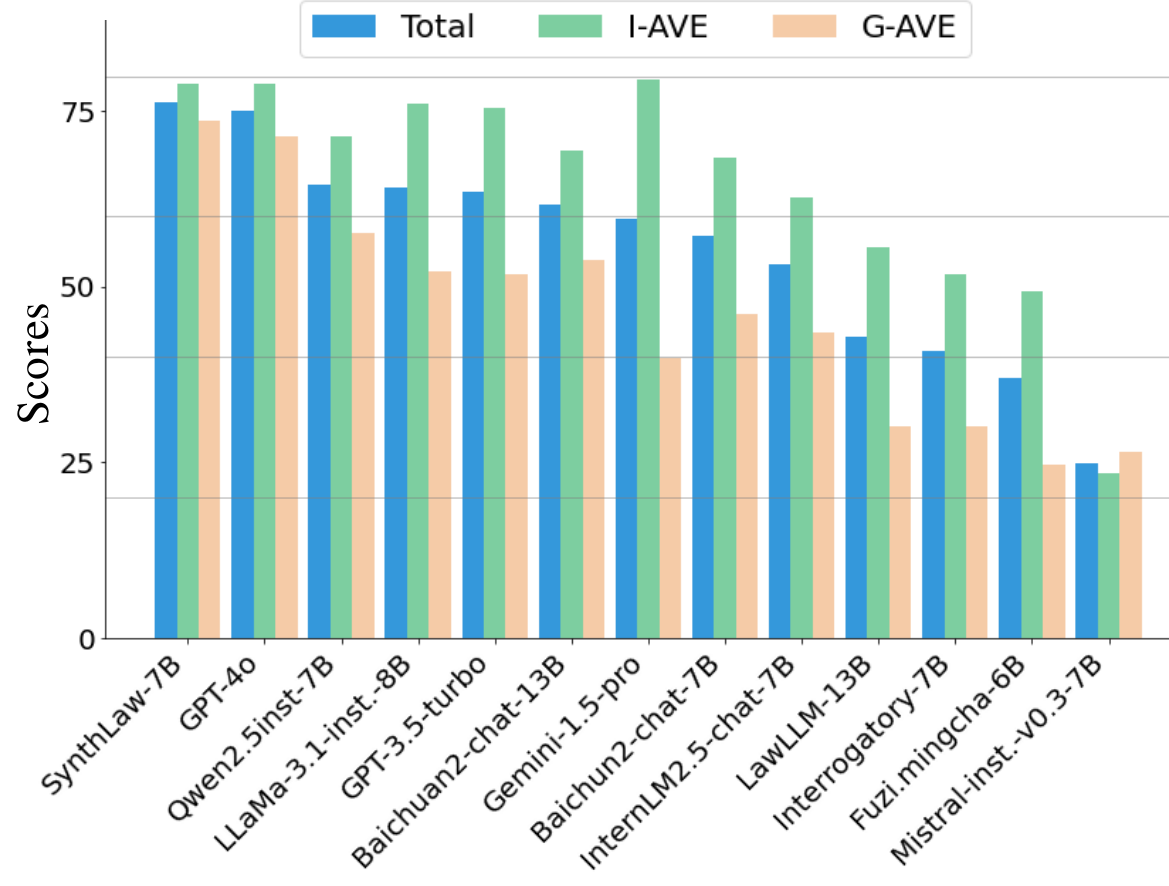}
    \caption{Comparative results of total performances, where G-AVE and I-AVE stand for goal evaluation and interaction evaluation average scores respectively.}
    \label{fig:total}
\end{figure}

\paragraph{Comparison on total performance.}
Total Performance aims to assess the average performance of the interaction and goal stages. As shown in Figure \ref{fig:total}, we can observe that our SynthLaw achieves the best performance in both the goal and total performance, even though it performs less effectively than existing closed-source LLMs in the interaction performance. 
Essentially, both stages are crucial for goal-oriented legal tasks: the former involves the complete collection of elements, while the latter focuses on transforming those elements into the final task output. This new intensive interaction scene is a necessary step toward achieving true legal intelligence. 
The experiment shows that our model effectively bridges these two stages.

\subsection{More Analysis}

\paragraph{Analysis of client behavioral consistency.}
To demonstrate the effectiveness of MILE Benchmark, we first analyze the client's consistent behavior across interactions with four LLM-driven lawyers.
Both GPT-4o and humans rate the behavior on a scale of 1 to 10.
Table \ref{tab:consistency} shows all the scores are high and stable across interactions with different lawyers, indicating that the client's behavior is reliable and consistent. 
% The Client, in a non-play role, presents various legal needs in a unique style.
This experiment validates the reliability and effectiveness of our multi-agent system, laying a solid foundation for assessing LLMs’ performance in interactive legal scenarios.

\begin{table}[]
\centering
\resizebox{0.48\textwidth}{!}{
\begin{tabular}{@{}ccccc@{}}
\toprule                       & GPT-4o & LawLLM & Qwen2.5 & SynthLaw \\ \midrule
\multicolumn{1}{c|}{GPT-4o} & 87.88  & 86.55  & 85.82 & 87.95 \\
\multicolumn{1}{c|}{Human}  & 87     & 81     & 84    & 87    \\ \bottomrule
\end{tabular}}
\caption{Client behavior consistency with various LLM-driven Lawyers under Human and GPT-4o evaluation.}
\label{tab:consistency}
\end{table}

\paragraph{Analysis of interaction with different clients.}
To further validate the robustness of our framework, we explore the performance of the lawyer models ( Initial LLM and SynthLaw) with different LLM-driven evaluation frameworks, where we set Qwen2.5-instruct 72B or GPT-4o as Client and Supervisor. 
As shown in Table \ref{tab:my-qwen}, the performance of the initial LLM shows little variation across different clients, indicating that our framework maintains relative stability under different LLMs. When both the client and supervisor are driven by a more powerful GPT-4o, the performance of the trained lawyer agent is gained even more.
This is because improved interactivity of the Client can enhance Lawyer's interactions. More importantly, SynthLaw achieves performance improvements under different clients, demonstrating that our method can effectively improve the model's ability for intensive interactions. 
In summary, the results not only validate the compatibility of the framework with different Clients and Supervisors, but also further validate our framework's effectiveness.
\begin{table}[]
\centering
\resizebox{0.9\columnwidth}{!}{%
\begin{tabular}{ccccc}
\hline
\multicolumn{1}{c|}{Model}        & INT   & PROF  & \multicolumn{1}{c|}{LOGI}  & AVE   \\ \hline
\rowcolor{gray!25}
\multicolumn{5}{c}{Client\space(GPT-4o)}                                                                \\
\multicolumn{1}{c|}{Initial LLM}  & 72.90 & 68.46 & \multicolumn{1}{c|}{72.73} & 71.36 \\
\multicolumn{1}{c|}{SynthLaw\space$_{\mathrm{7B} }$}        & 83.23 & 74.48 & \multicolumn{1}{c|}{79.22} & 78.97 \\
\rowcolor{gray!25}
\multicolumn{5}{c}{Client\space(Qwen2.5-instruct\space$_{\mathrm{72B} }$)}                                                               \\
\multicolumn{1}{c|}{Initial LLM}  & 72.46 & 67.59 & \multicolumn{1}{c|}{71.67} & 70.58 \\
\multicolumn{1}{c|}{SynthLaw\space$_{\mathrm{7B} }$} & 78.26 & 70.83 & \multicolumn{1}{c|}{74.64} & 74.58 \\ \hline
\end{tabular}%
}
\caption{Interaction with different Client Models, where INT, PROF, and LOGI are abbreviations of Interactivity, Professionalism, and Logicality respectively.}
\label{tab:my-qwen}
\end{table}

\paragraph{Analysis of different LLMs driven by MASER.}
We use the SynthLaw dataset generated by MASER to train three different initial models (Baichuan2-chat-7B, InternLM2.5-chat-7B and Qwen2.5-instruct-7B), resulting in three distinct SynthLaw models. The performances on interaction evaluation are shown in the table \ref{tab:vanilla-trained}, we can observe that across the three base models, SynthLaw improves significantly in all the metrics, particularly bringing a 28.1\% average performance boost to Internlm2.5-chat. 
This shows that our MASER can drive arbitrary LLMs, enabling them to perform intensive interactions in dynamic legal scenarios. The goal evaluation is provided in Appendix \ref{sec:appendix: dataset_gmaser}.

\section{Related Work}

\paragraph{Legal LLM.}
Legal-domain LLMs have achieved astounding performance on legal tasks, such as legal information extraction~\cite{bommarito2018lexnlp}, case retrieval~\cite{ma2021lecard}, judgment prediction~\cite{huang2021dependency}, 
which offer broad applications that benefit different groups of the population.
Initial progress \cite{lawyerllama,yue2024lawllm,deng2023syllogistic} has been made by fine-tuning general LLMs to utilize legal knowledge for different legal tasks.
Specifically, Lawyer-LLaMa \cite{lawyerllama} and Interrogatory inject domain knowledge during continuous training. 
Fuzimingcha trained on a vast corpus of unsupervised Chinese legal texts and supervised judicial fine-tuning data.
LawLLM \cite{yue2023disc,yue2024lawllm} introduces legal retrieval capability to enhance factuality.
Previous approaches focused on static tasks, ignoring the dynamic properties of real-world legal tasks. To fill this gap, this study places emphasis on intensive legal interactions.

\paragraph{Role-playing Agent.}
The advancement of LLM-powered agents has greatly improved complex task resolution through anthropomorphic actions \cite{park2023generative,fan2024ai,yue2024synergistic}.
By mimicking human sense and vivid performance, role-playing agents present great potential in various fields \cite{mou2024agentsense,gao2024fine,lyu2024human,liu2024ai}.
However, in the legal field, the limited expertise of LLMs makes it challenging for existing role-playing methods \cite{xie2024can,jiang2024evaluating} to simulate legal attributes in multi-agent scenarios (\textit{e.g.}, clients and legal providers). This requires not only establishing legal attribute correspondences between different agents but also ensuring consistency in their profile and behavior under intensive interactions. To this end, we propose the MASER framework.

\begin{table}[]
\centering
\resizebox{\columnwidth}{!}{%
\begin{tabular}{c|ccc|c}
\hline
Model                                          & INT            & PROF           & LOGI           & AVE            \\ \hline
Baichuan2-chat                                 & 72.58          & 65.33          & 70.37          & 69.42          \\
\multirow{2}{*}{SynthLaw $_{\mathrm{Baic}}$}   & \textbf{83.77} & \textbf{75.26} & \textbf{78.96} & \textbf{79.33} \\
                                               & (15.4\%$\uparrow$)             & (15.2\%$\uparrow$)             & (12.2\%$\uparrow$)             & (14.3\%$\uparrow$)       \\ \hdashline
Internlm2.5                                    & 64.35          & 60.18          & 63.81          & 62.78          \\
\multirow{2}{*}{SynthLaw $_{\mathrm{Intern}}$} & \textbf{84.93} & \textbf{76.19} & \textbf{80.07} & \textbf{80.40} \\
                                               & (32.0\%$\uparrow$)             & (26.6\%$\uparrow$)             & (25.5\%$\uparrow$)             & (28.1\%$\uparrow$)       \\ \hdashline
Qwen2.5-inst.                                  & 72.90          & 68.46          & 72.73          & 71.36          \\
\multirow{2}{*}{SynthLaw $_{\mathrm{Qwen}}$}   & \textbf{83.23} & \textbf{74.48} & \textbf{79.22} & \textbf{78.97} \\
                                               & (14.2\%$\uparrow$)             & (8.8\%$\uparrow$)             & (8.92\%$\uparrow$)             & (10.7\%$\uparrow$)       \\ \hline
\end{tabular}}
\caption{Performances of initial LLMs and their corresponding trained versions in the Interaction evaluation.}
\label{tab:vanilla-trained}
\end{table}

\section{Conclusion}
In this paper,
we introduce the Multi-agent Legal Simulation Driver (MASER), a legal-specific simulator that serves as data-generation engine, empowering arbitrary LLMs with intensive interaction capabilities.
In MASER, we establish consistency in the legal attributes among roles using real legal case sources, 
and introduce a supervisory mechanism to align the characters and behaviors during interactions, which enables high-quality and sentence-level aligned legal interaction data.
In addition, an interactive legal benchmark, Multi-Stage Interactive Legal Evaluation (MILE), is proposed to evaluate the capacity of LLMs as lawyers in performing legal tasks (\textit{i.e.}, complaint drafting ) within dynamic scenarios. 
The experimental results 
demonstrate the effectiveness of our MASER. Our framework can extend more complex domain scenarios, bridging the gap between intensive interaction and achieving special objectives.

\section*{Limitations}
In this paper, we take a first step forward from static to a dynamic, interactive, legal task.
In our multi-agent simulation framework, the ultimate legal task is defined as the generation of indictments. Although we have established indictment generation across various scenarios, dynamic legal contexts extend beyond this scope. In the future, we aim to expand our framework to encompass diverse legal scenarios, such as courtroom proceedings and legal consultations. 

\section*{Acknowledgments}
The work is supported by National Key R\&D Program of China (Grant Nos. 2023YFF1204800), National Natural Science Foundation of China (Grant Nos. 62176058) and Shaanxi Province Social Science Fund Projec (Grant Nos. 2023XWT04). The project’s computational resources are supported by CFFF platform of Fudan University.
\bibliography{custom}

\appendix
\section{Role Presetting Details}
\begin{table*}[t!]
\begin{center}
\resizebox{\textwidth}{!}{%
\begin{tcolorbox}
[colback=black!5!white,colframe=gray!15!gray,width=\textwidth,title={Legal Agenda for Complaint Drafting.}]
\textbf{[Agenda 1]}: Clint's Basic Information: Name, gender, date of birth, ethnicity, and address.

\textbf{[Agenda 2]}: Defendant's Basic Information.

\textbf{[Agenda 3]}: Basic Case Information: The time, place, full details of the event, and key points of contention.

\textbf{[Agenda 4]}: Clint's Claims: What outcome is desired from this lawsuit, such as compensation amount or specific actions
requested from the other party.

\textbf{[Agenda 5]}: Litigation Costs: Whether the plaintiff seeks to have the defendant cover the litigation costs.

\textbf{[Agenda 6]}: Contracts, agreements, receipts, physical evidence, witness information or testimony, recordings, expert
reports, videos, etc

\textbf{[Agenda 7]}: Any adverse evidence from the defendant or other relevant information.
\end{tcolorbox}}
\end{center}
    \caption{Legal Agenda setting by expert.}
    \label{tab:examplse_agenda}
\end{table*}

\subsection{Judgement Document Extraction}
\label{sec:appendix: dataset_jd}

\begin{figure*}[t]
    \centering
\includegraphics[width=0.9\linewidth]{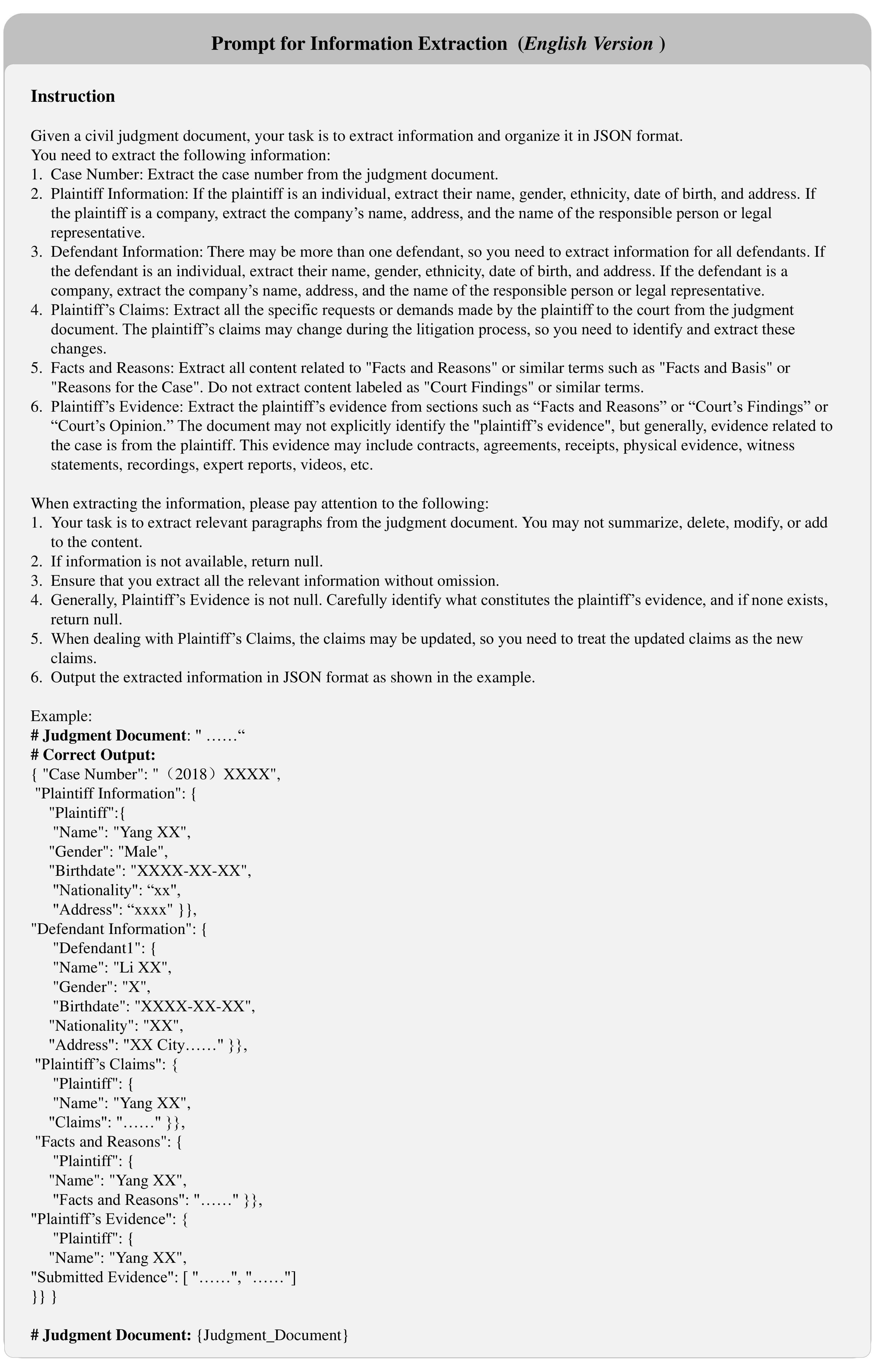}
    \caption{Prompt for Information Extraction(\textit{English Version})}
    \label{fig:p9}
    % \vspace{-3mm}
\end{figure*}

Judicial Document is the record of the court's proceedings and outcomes. It serves as the carrier of the results of litigation activities and is the sole evidence by which the court determines and allocates the substantive rights and obligations of the parties involved. 
It is characterized by its complete structure, comprehensive elements, and rigorous logic.
Due to the lack of legal dynamic data, we skillfully utilize such legal documents to develop interactive scenarios. 
We extract the desired legal elements from the documents and then configure them into agents to drive their knowledge and behavior. 
Since the documents contain the complete evolution of events, this way ensures logical and realistic interactions between agents.
Specifically, we extracted the following seven elements by utilizing an extraction model (GPT-4o):

\begin{itemize}
    \item \textbf{Plaintiff information} includes name, gender, nationality, birthdate, and address.
    \item \textbf{Defendant information} has two categories: individuals include name, gender, nationality, birthdate, and address; Companies include the company's name, address, and the name of the responsible person or legal representative.
    \item \textbf{Claim} is the demand or requests made by the plaintiff to the court, including litigation fees.
    \item \textbf{Case detail} details the events between the plaintiff and the defendant.
    \item \textbf{Evidence} is material submitted by the plaintiffs in support of their claims.
    \item \textbf{Case analysis} is a detailed and authoritative analysis of a case by the court using facts, evidence and applicable law. 
    \item \textbf{Legal provisions} are the exact legal rules given by the court that apply to the case.
\end{itemize}
    These categories of elements are assigned to the appropriate agents within our framework. 
    Extraction prompt refers to Figure \ref{fig:p9}.

\subsection{Legal Agenda}
The legal agenda provides legal service providers with a standardized operational framework, reducing unnecessary disputes and uncertainties. Through systematic legal rules, legal service providers are able to address legal issues more efficiently, thereby improving the quality of their services. Understanding and adhering to legal rules is at the core of their professional responsibilities.
In complaint drafting services, legal agenda guide lawyers to understand the user's claims and gather accurate information. As shown in Figure \ref{tab:examplse_agenda}, it involves the following key process: client information, defendant information, case fact,  controversy, appeal and applicable evidence.

\subsection{Personality  Modeling}
\label{sec:appendix: dataset_pt}
\begin{figure*}[t]
    \centering
\includegraphics[width=0.91\linewidth]{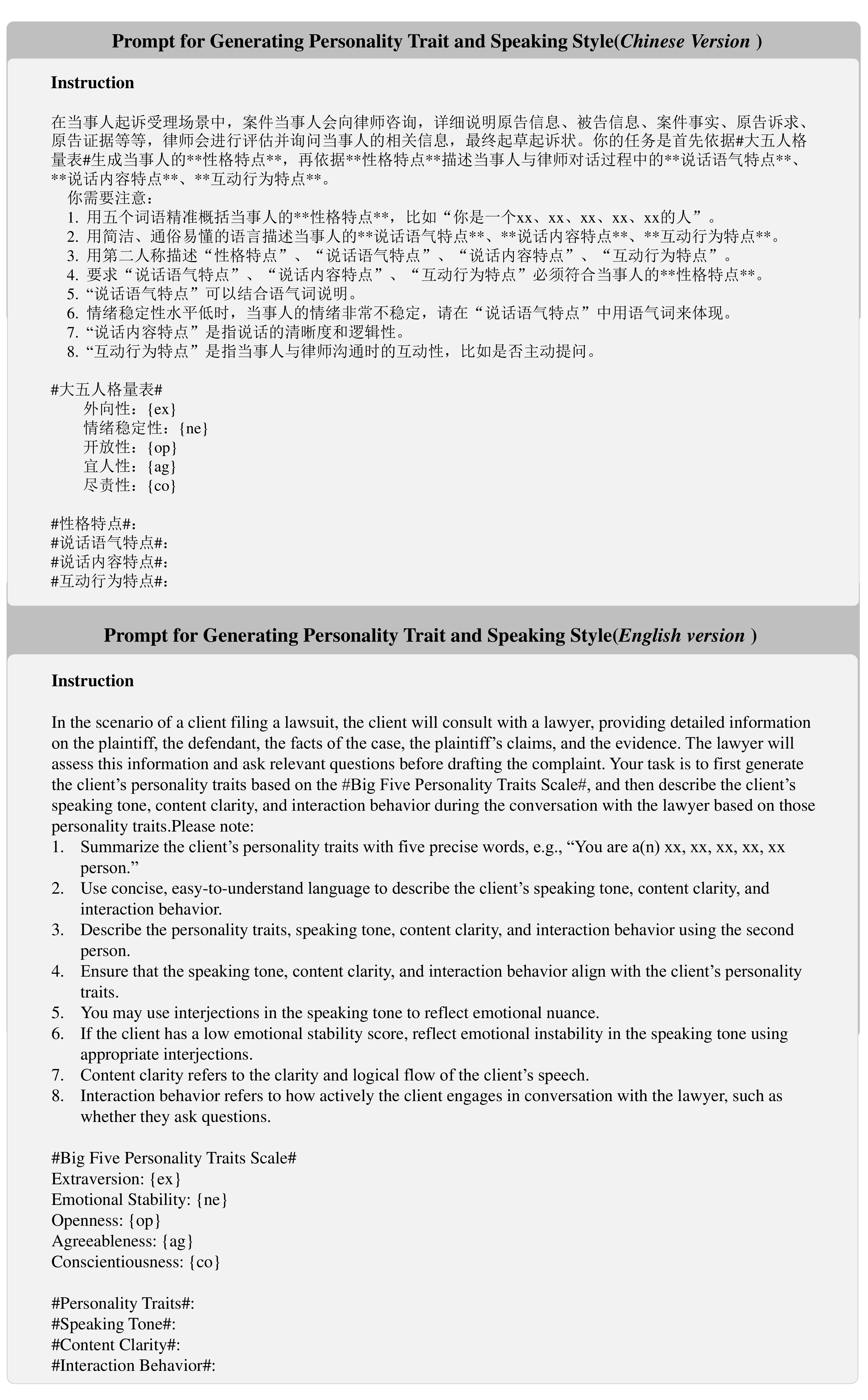}
    \caption{Prompt for Generating Personality Trait and Speaking Style}
    \label{fig:p19}
\end{figure*}

\paragraph{Big Five Personality Traits.} 
The client's diversity facilitates enhancing the diversity and generalization of the data.
We construct multi-level user characteristics based on the Big Five Personality Traits theory, which has five dimensions: 
encompasses five dimensions: extraversion, emotional stability, openness, agreeableness, and conscientiousness. Exiting studies \cite{sun2024identity,tseng2024two} have demonstrated that this theory can assist LLMs to understand better the roles played. 
In our implementation,
we frist divide each dimension of the theory into three levels (high, medium, low) and randomly combine them to form five traits. To enhance the distinctiveness of the character portrayals, the distribution ratio of high, medium, and low levels is set to 2:1:2. Additionally, considering that the individuals involved in the case are typically inclined to anxiety, we increase the probability of emotional stability being at high levels. Based on these traits, we prompt the GPT-4o to generate a brief character's personality, and further generate the character's speaking style and interaction behavior, where speaking style consists
of logic, clarity and tone. The prompt is shown in Figure \ref{fig:p19}.
\paragraph{Legal Sense.}
Five levels of legal sense are manually generated by legal experts to more realistically simulate the parties in the interaction scenarios. The definitions from low to high are as follows: 
\begin{itemize}
    \item \textbf{Level 1.} Completely lacks legal knowledge and is unable to use any legal-related terminology, such as "rights" or "obligations." Responses focus primarily on the straightforward description of events.
    \item \textbf{Level 2.} Has basic legal awareness and knows simple legal terms such as "litigation" or "breach of contract," but does not fully understand their specific meanings. Responses attempt to engage with legal aspects, though there may be inappropriate usage of terms, with an emphasis still on narrating the concrete situation.
    \item \textbf{Level 3.} Possesses foundational legal knowledge and can correctly use everyday legal terms and expressions such as "contract terms" or "litigation." Responses incorporate legal terminology in describing the situation.
    \item \textbf{Level 4.} Familiar with basic legal terminology and able to accurately use more complex legal terms and concepts, such as "right to litigate" or "enforcement of judgment."
    \item \textbf{Level 5.} Highly proficient in legal knowledge, familiar with fundamental legal provisions, and able to describe legal issues in detail. Additionally, can propose legal strategies or defense points that may be beneficial to the case.
\end{itemize}

\section{MILE Benchmark Detail}

\subsection{Interaction Evaluation}
\label{sec:appendix: interaction_evaluation}
Unlike a direct assessment of the entire interaction history, we use a fine-grained interaction assessment.
In our interactive scenarios, the information and the logic of the previous turn are typically associated with next turn. For example, when asked personal information, the client may miss some of the details, and the lawyer should ask follow-up questions to clarify those missing details in the next turn. Therefore, we adopt a two-turn window as our fine-grained evaluations, which can keep a trade-off between evaluation accuracy and evaluation costs. 

In our evaluation, GPT4-o serves as a referee and performs the evaluation by providing a rating score from 1 to 10 for each of the following three criteria: \textit{interactivity}, \textit{professionality}, and \textit{logicality}.
\begin{itemize}
    \item Interactivity: the model should proactively participate in the dialogue, answering and asking questions that would advance the discussion and clarify any vagueness.

    \item Professionality: the model should correctly use legal terms, cite relevant laws and precedents, as well as offer professional strategies to the client.

    \item Logicality: the model should sustain logical conversations without repeating any of the previously discussed topics.
\end{itemize}

\begin{figure*}[t]
    \centering
\includegraphics[width=0.9\linewidth]{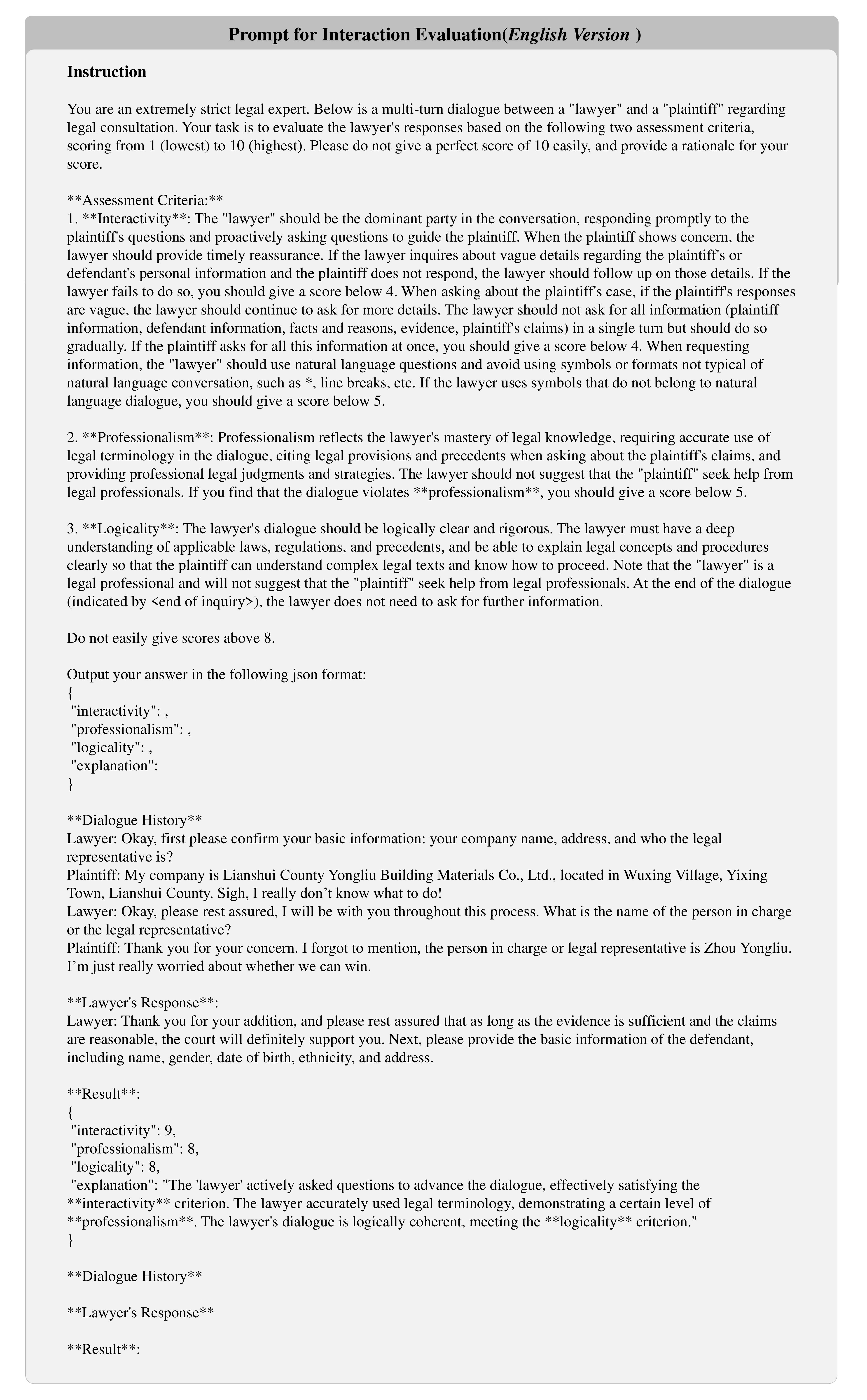}
    \caption{Prompt for Interaction Evaluation(\textit{English Version})}
    \label{fig:p12}
\end{figure*}

The prompt for GPT-4o is provider as Figure \ref{fig:p12}. Scores for each metric are then obtained by calculating the average score for each window.

\begin{figure*}[t]
    \centering
\includegraphics[width=0.8\textwidth]{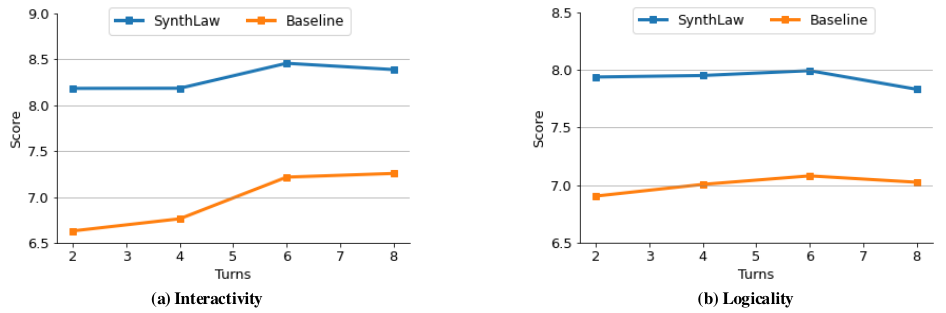}
    \caption{The scores (\textit{Interactivity} and \textit{Logicality}) over different  turn numbers on interaction evaluation, where the baseline is Qwen2.5-instruct-7B.}
    \label{fig:Three_LINE}
    % \vspace{-3mm}
\end{figure*}
\subsection{Goal Evaluation}
\begin{figure*}[t]
    \centering
\includegraphics[width=0.93\linewidth]{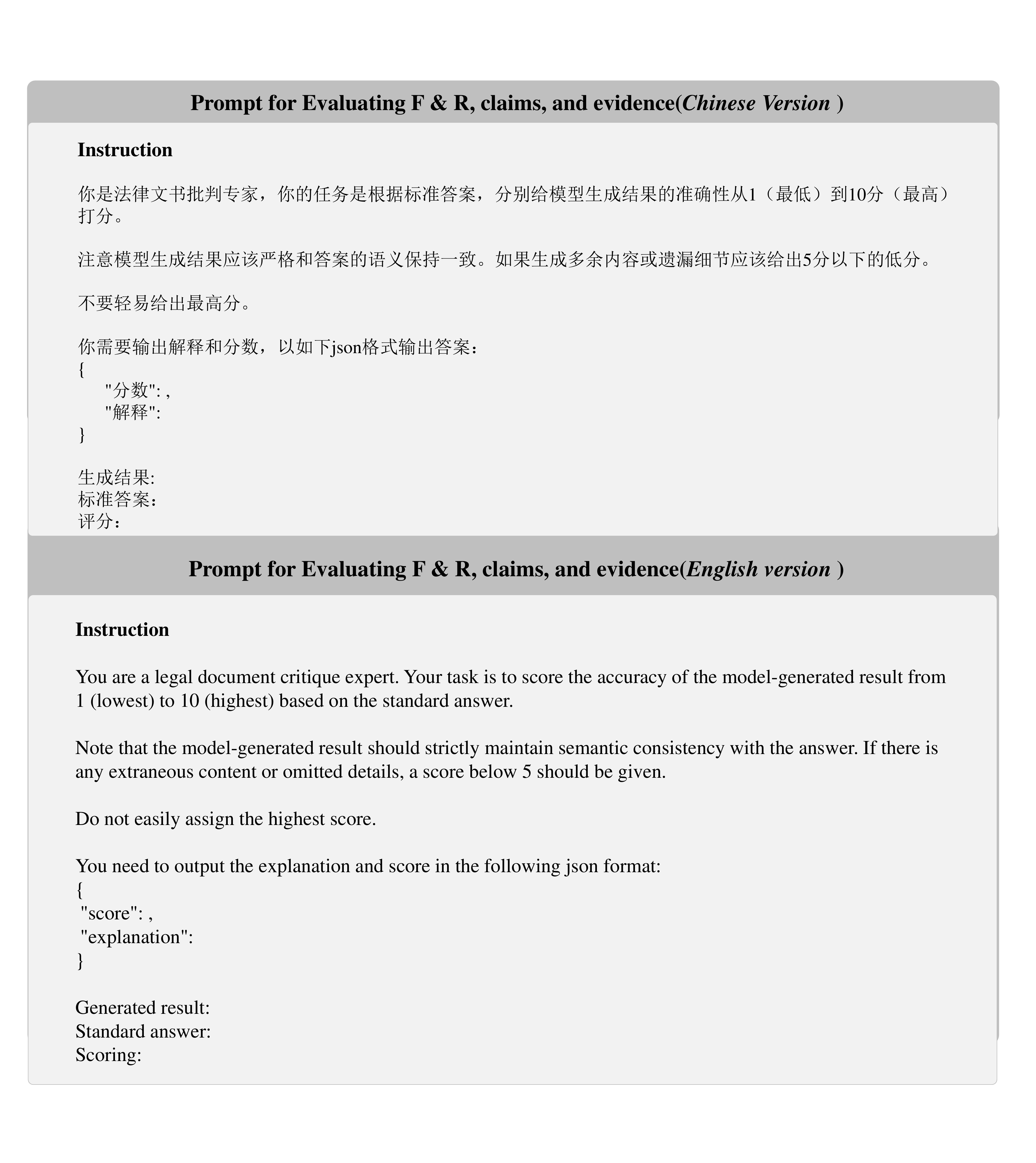}
    \caption{Prompt for Evaluating F \& R, claims, and evidence}
    \label{fig:p15}
\end{figure*}

The goal evaluates the quality of complaints quality from local and global perspectives.
For each goal evaluation sample, a ground
truth is provided to reduce potential biases during the assessment phase.
\begin{table}[t]
\centering
\resizebox{0.48\textwidth}{!}{
\begin{tabular}{@{}lllll@{}}
\toprule
Legal Attributes       & PROF  & LOGI  & INT   & AVE            \\ \midrule
Contracts Dispute      & \cellcolor[HTML]{5eb95f}73.96 & \cellcolor[HTML]{5eb95f}78.14 & \cellcolor[HTML]{a5daa6}81.58 & \cellcolor[HTML]{5eb95f}\textbf{77.89} \\
Economic Dispute       & 71.26 & 75.58 & 79.60 & 75.48          \\
Property Dispute       & 73.24 & 76.36 & 79.49 & 76.36          \\
Marital \& Inheritance & \cellcolor[HTML]{a5daa6}73.82 & \cellcolor[HTML]{a5daa6}76.91 & \cellcolor[HTML]{5eb95f}82.86 & \cellcolor[HTML]{a5daa6}77.86          \\
Labor Dispute          & \cellcolor[HTML]{E66666}66.62 & \cellcolor[HTML]{E66666}68.65 & \cellcolor[HTML]{E66666}70.50 & \cellcolor[HTML]{E66666}68.59          \\
Personality Rights     & \cellcolor[HTML]{FFB6C1}71.01 & \cellcolor[HTML]{FFB6C1}74.89 & \cellcolor[HTML]{FFB6C1}78.52 & \cellcolor[HTML]{FFB6C1}74.85          \\
Maritime Dispute       & 73.50 & 76.83 & \cellcolor[HTML]{a5daa6}81.58 & 77.30          \\
Rights Infringement    & 73.64 & 75.45 & 81.27 & 76.79          \\
Intellectual Property  & 73.02 & 76.59 & 79.30 & 76.30          \\ \bottomrule
\end{tabular}}
\caption{Performances on different legal attributes of interaction evaluation, where INT, PROF, and LOGI are abbreviations of Interactivity, Professionalism, and Logicality respectively. Darker (best) to lighter green marks the best of the top two results, while darker (worst) to lighter red marks the worst of the top two results.}
\label{tab:Attributes-interaction}
\end{table}
\paragraph{Local Evaluation.} 
Since the complaint presents a high degree of structure, we evaluate each part of the complaint: client information (CLI), defendant information (DEF), facts \& reasons (F \& R), claims (CLA) and evidence (EVID). We follow two guidelines, for short-form generation (\textit{e.g.}, CLI) we calculate the accuracy directly by matching. For long-form generation  (\textit{e.g.}, CLA), we use GPT-4o to calculate the score based on the semantic similarity with ground truth. The details are as follows:
% We compare the extracted client information, defendant information, facts, reason, claims, and evidence with their counterparts in the ground truth.

\begin{itemize}
    \item \textit{CLI and DEF}. we examine if they are identical to the ground truth and calculate an accuracy as the final score.

    \item \textit{F \& R, CLA and EVID}. We prompt GPT-4o to rate from 1 (lowest) to 10 (highest). The prompt is shown as Figure \ref{fig:p15}.
\end{itemize}

\begin{figure*}[t]
    \centering
\includegraphics[width=0.9\linewidth]{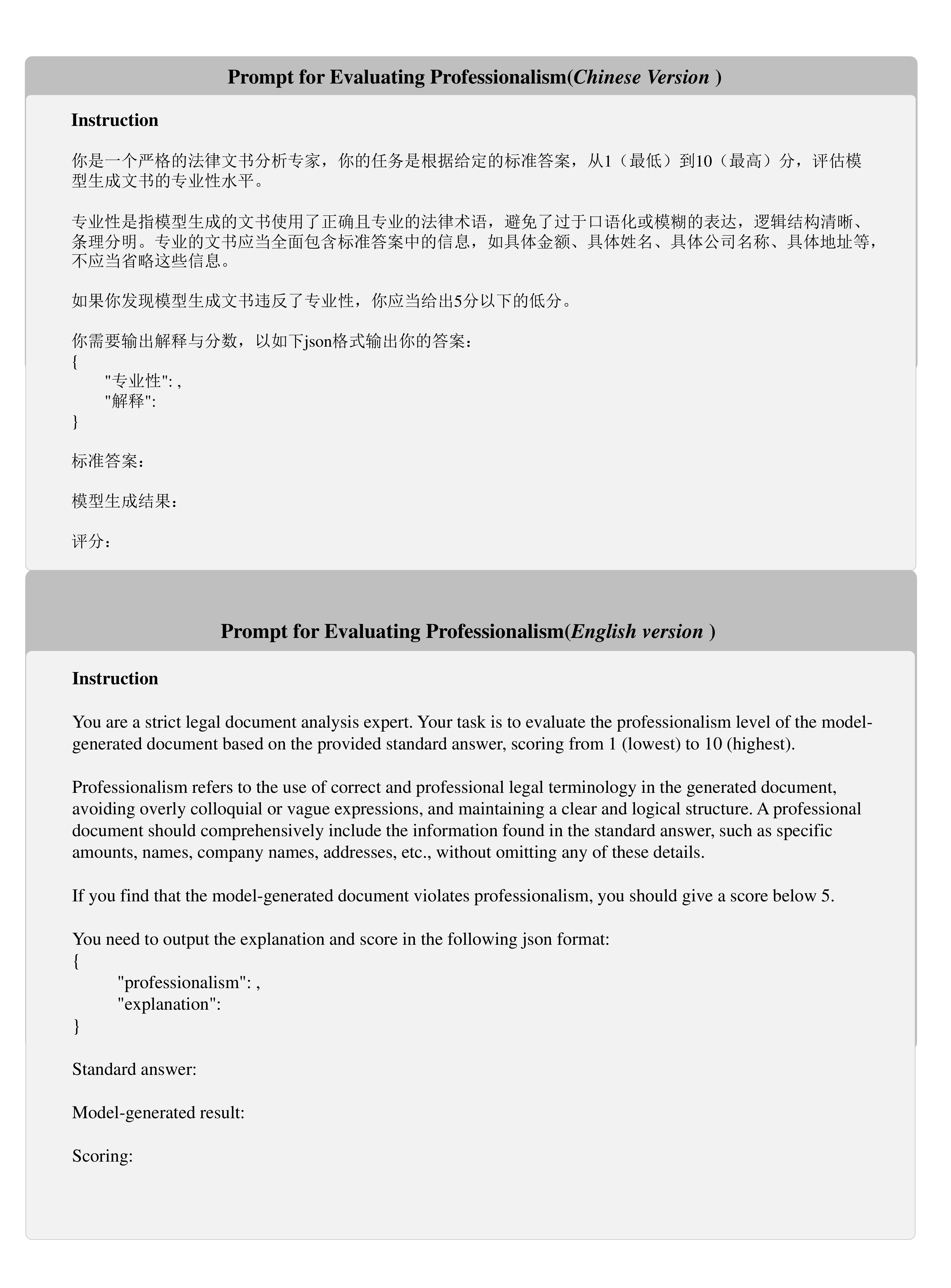}
    \caption{Prompt for Evaluating Professionalism}
    \label{fig:p13}
\end{figure*}

\begin{figure*}[t]
    \centering
\includegraphics[width=0.8\linewidth]{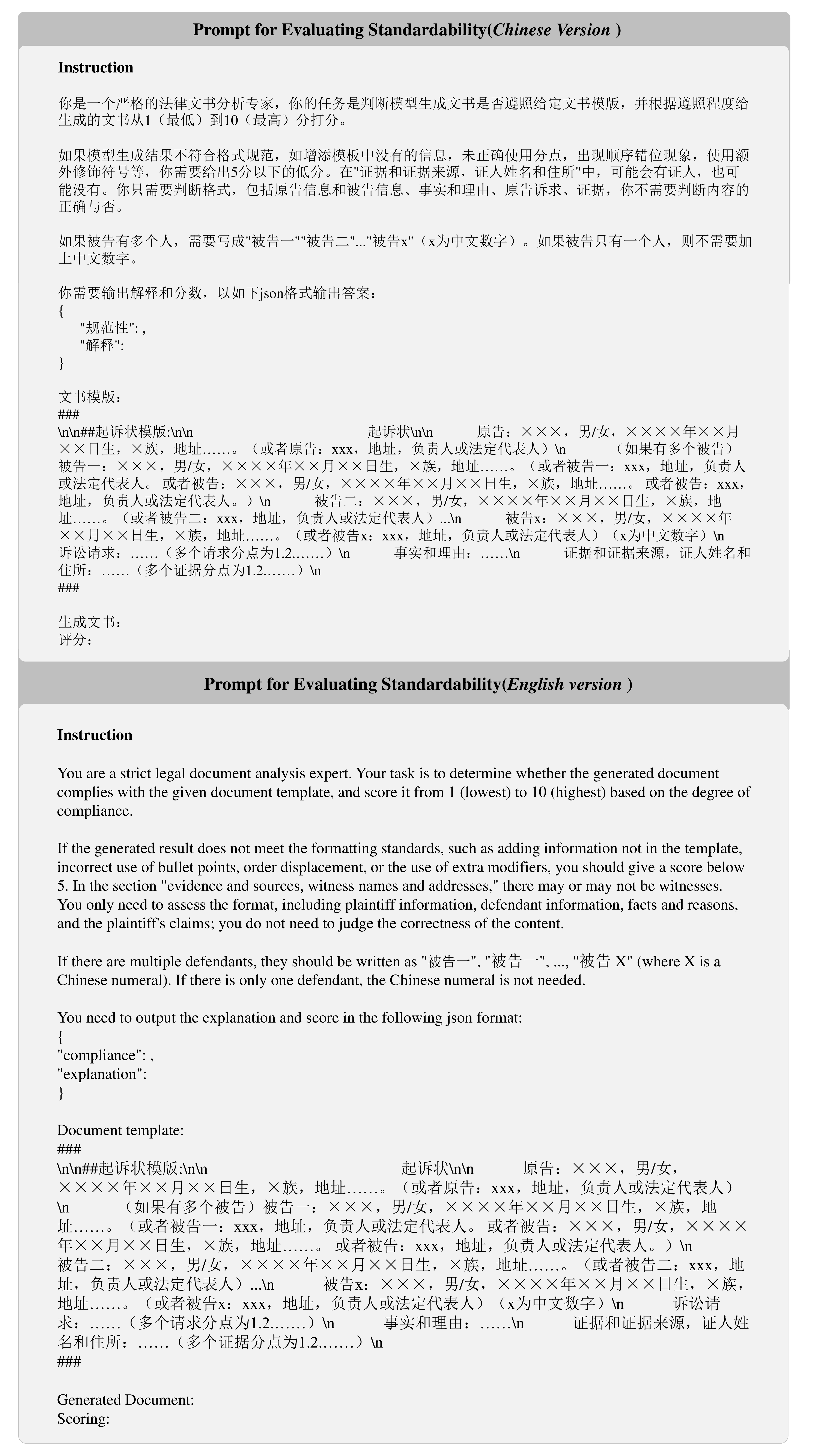}
    \caption{Prompt for Evaluating Standardability}
    \label{fig:p14}
\end{figure*}

\paragraph{Global Evaluation.} 
Besides the accuracy assessment described above, Global Evaluation assesses the overall \textit{Standardability} (whether the document follows a given template) and \textit{Professionalism} (whether correct legal language is used) of the complaint. 
We also prompt GPT-4o to rate from 1 (lowest) to 10 (highest). 
\begin{itemize}
    \item Standardability: follows a given document template and focus on format, not specific content.
    The prompt is shown as Figure \ref{fig:p14}.

    \item Professionalism: refers to the use of correct and professional legal terminology in the generated document, avoiding overly colloquial or vague expressions, and maintaining a clear and logical structure. The prompt is shown as Figure \ref{fig:p13}.
\end{itemize}

\label{sec:appendix: goal_evaluation}

\section{Implementation Details}
\label{sec:appendix: setting_details}
\paragraph{Training Detail.} 
We use  Qwen2.5-instruct-7B \cite{yang2024qwen2} as our initial model. 
We use 8*RTX 4090 GPUs with 24GB memory to conduct the LoRA method \cite{hu2021lora}.
Our models are trained for 8 epochs with a batch size of 32, and a peak learning rate of 2e-4. We set the maximum token length to be 2,048. Multi-GPU distributed training is performed using DeepSpeed Stage 2 \cite{rasley2020deepspeed}, with training precision Bfloat16 enabled.

\paragraph{Evaluation Details.}
In implementation, we use GPT-4o\footnote{gpt-4o-2024-08-06}\cite{achiam2023gpt} to drive client and supervisor in MILE Benchmark.
For adapted baselines, we speed up inference using vllm \cite{kwon2023efficient}. Greedy decoding was used across the evaluations. We run evaluations using 1-2 V100 GPUs with 32GB memory.
\section{Additional Experiments}

\begin{table*}[!t]
\centering
\resizebox{0.96\textwidth}{!}{%
\begin{tabular}{l|ccccccc|c}
\hline
\multirow{2}{*}{Model} & \multicolumn{5}{c}{Local}              & \multicolumn{2}{c|}{Global} & \multirow{2}{*}{AVE} \\
                                  & CLI   & DEF   & F \& R & CLA   & EVID  & STA          & PROF         &                      \\ \hline
Baichuan2-chat                 & 79.37 & 78.44 & 34.21     & 47.97 & 29.39 & 55.73        & 44.46        & 46.20  \\
\multirow{2}{*}{SynthLaw $_{\mathrm{Baic}}$}             & \textbf{79.48} & \textbf{85.02} & \textbf{47.97}  & \textbf{64.43} & \textbf{31.70} & \textbf{87.27}           & \textbf{58.59}        & \textbf{60.85}               \\
&(0.14\%$\uparrow$) & (8.39\%$\uparrow$) &(40.22\%$\uparrow$) &(34.31\%$\uparrow$) &(7.86\%$\uparrow$) & (56.59\%$\uparrow$) & (31.78\%$\uparrow$) & (31.71\%$\uparrow$) 
\\ \hdashline
Internlm2.5                  & 62.77 & 64.28 & 35.74  & 44.62 & 31.03 & 50.97        & 44.21        & 43.51                \\
\multirow{2}{*}{SynthLaw $_{\mathrm{Intern}}$}          & \textbf{96.59} & \textbf{96.72} & \textbf{48.15}  & \textbf{69.80} & \textbf{31.80} & \textbf{87.81}        & \textbf{64.44}        & \textbf{65.82}                \\ 
& (53.88\%$\uparrow$) &  (50.47\%$\uparrow$) &  (34.72\%$\uparrow$) &  (56.43\%$\uparrow$) &  (2.48\%$\uparrow$) &  (72.28\%$\uparrow$) &  (45.76\%$\uparrow$) & (51.28\%$\uparrow$)
\\\hdashline
Qwen2.5-inst.                     & 86.90 & 92.05   & 37.81  & 55.66 & 34.23    & 41.50        & 56.09        & 57.75                \\
\multirow{2}{*}{SynthLaw $_{\mathrm{Qwen}}$}               & \textbf{90.20} & \textbf{96.15} & \textbf{54.20}  & \textbf{70.00} & \textbf{54.60} & \textbf{91.74}        & \textbf{59.05}        & \textbf{73.71}                \\
&(3.80\%$\uparrow$) & (4.45\%$\uparrow$) &(43.35\%$\uparrow$) &(25.76\%$\uparrow$) &(59.51\%$\uparrow$) & (121.06\%$\uparrow$) & (5.28\%$\uparrow$) & (27.64\%$\uparrow$) \\ \hline
\end{tabular}}
\caption{Performances of initial LLMs and their corresponding trained versions in the goal evaluation. The \textbf{bold} numbers represent the best results.}
\label{tab:Different LLMs driven by MASER in Goal stage}
\end{table*}

\subsection{Performances on different legal Attributes of MILE}
We explore SynthLaw's performance in completing complaints with different legal attributes, including intellectual property, tort liability, maritime dispute, personality rights, labor dispute, marital \& inheritance, economic dispute, and contracts dispute. As shown in table \ref{tab:Attributes-interaction}, among all legal attributes, most of legal attributes share similar scores, with the topic of labor disputes scoring lower than other topics, even with topic-related knowledge provided. This may due to insufficient pre-training of the base model on the topic of labor disputes.
The experiment highlights these discrepancies, offering valuable insights to guide future works in a more nuanced manner, particularly in addressing the specific types of disputes.

\begin{table}[]
\centering
\setlength{\tabcolsep}{4mm}{
\resizebox{0.4\textwidth}{!}{%
\begin{tabular}{@{}llll@{}}
\toprule
Model    & Ave  & Max & Min \\ \midrule
Baichuan2-chat & 9.94& 10 & 2 \\
SynthLaw $_{\mathrm{Baic}}$ & 9.16& 15 & 3 \\\hdashline
Internlm2.5 & 4.29 & 10 & 1\\
SynthLaw $_{\mathrm{Intern}}$ & 8.24 & 15 & 3\\\hdashline
Qwen2.5-inst. & 6.24 & 10  & 1   \\
SynthLaw $_{\mathrm{Qwen}}$ & 8.45 & 15  & 3   \\ \bottomrule
\end{tabular}%
}}
\caption{Ave, max and min number of interaction turns for our SynthLaw and baseline models}
\label{tab:tun}
\end{table}
\subsection{Interaction score over different turns}
To figure out the interaction performance, we show the Interactivity and Logicality of models on the previous 8 rounds. In our implementation, the initial LLM is Qwen2.5-instruct 7B.
As shown in Figure \ref{fig:Three_LINE}, 
we observed that our model outperformed the baseline model across all metrics in every round. Additionally, we note that the scores in the first six rounds were comparatively lower, as these rounds involved the collection of personal information. This process poses greater challenges to the model's interaction capabilities due to the user's distracting behaviors (\textit{e.g.}, missing details). Nevertheless, the trained model exhibits significant performance improvement and maintained relative stability, further demonstrating that our framework effectively enhances the model's ability to adapt flexibly to the specified legal agenda.

\label{sec:appendix: Interaction_score_over_turns}

\subsection{Performances on different LLMs driven by MASER}
We utilize the SynthLaw dataset generated by our framework to train three different models(i.e., Baichuan2-chat-7B, InternLM2.5-chat-7B, and Qwen2.5-instruct-7B) into three different SynthLaw models(i.e., SynthLaw-Baic, SynthLaw-Intern, and SynthLaw-Qwen). We then conduct experiments to see these models' performance in Goal stage. As shown in table \ref{tab:Different LLMs driven by MASER in Goal stage}, for all three of the base models, their performance has greatly improved after being trained under our MASER framework. This has demonstrated that our framework's effect could generalize to different base models, improving their ability to extract correct information from the interaction and follow a given format.

\label{sec:appendix: dataset_gmaser}
\begin{figure*}[t]
    \centering
\includegraphics[width=1\linewidth]{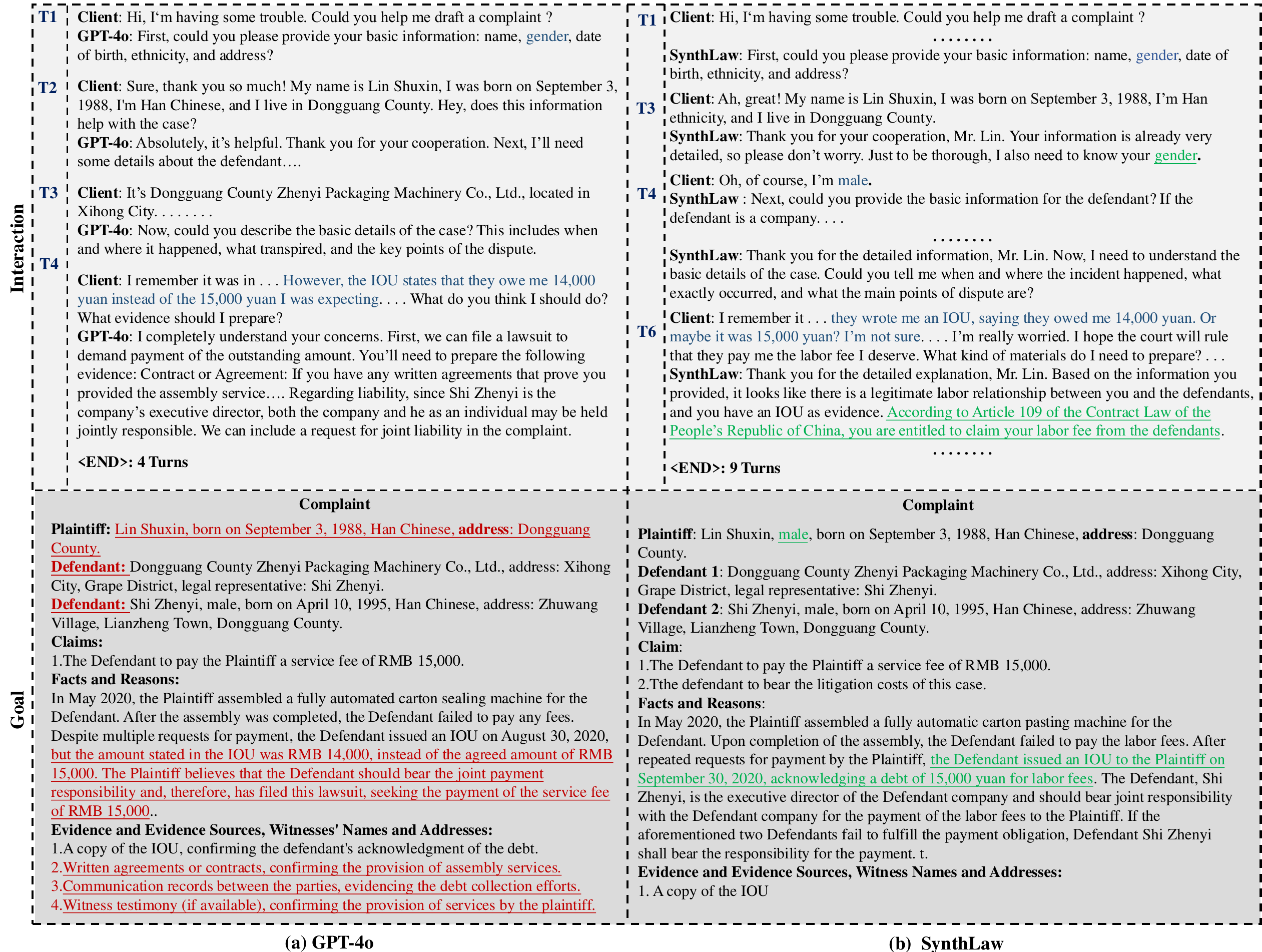}    \caption{Qualitative result of our SynthLaw 7B against GPT-4o on MILE benchmark. T$i$ denotes the $i$-th interaction turn. Green underlines highlight responses, while red underlines denote incomplete or incorrect responses.}
    \label{fig:case}
\end{figure*}
\subsection{Interaction turn numbers on different LLMs}
We count the average, maximum, and minimum number of interactions for different baseline models and their trained versions in evaluation.  
From Table \ref{tab:tun}, we observe that the trained models have longer interaction numbers than their corresponding initial models, especially for InternLM2.5-chat-7B and Qwen2.5-instruct-7B.  
A higher number of interactions indicates that the model actively seeks detailed information from the user to comprehensively address their needs. In contrast, fewer turns may result in the omission of critical details, thereby limiting the model's ability to comprehend the user's intent.  
Notably, the average number of SynthLaw $_{\mathrm{Baichuan}}$ is slightly lower than Baichuan2-Chat, yet it achieves higher scores on the Goal and Interaction evaluation.  
Baichuan2-chat tends to generate repeated greetings rather than ending the conversation in time.
Experimental results show that our method significantly enhances the model's ability to engage in dense interactions, further proving the effectiveness of our approach.
% \subsection{Goal Evaluation}
% \label{sec:appendix: dataset_ge}
\subsection{Case Study}
To present the performance generated by the proposed framework, we conduct a qualitative study in MLIK with SynthLaw 7B and GPT-4o acting as lawyers, where SynthLaw 7B is
initialized from Qwen2.5-instruct 7B.
As illustrated in Figure \ref{fig:case}, during the interaction phase, when the Client omits to provide gender information, GPT-4o ignores this and proceeds with the agenda (T2).
In contrast, SynthLaw successfully followed up to inquire about the missing gender information (T4). when users include vague expressions involving a monetary amount of 1500 in the case description, GPT-4o neither confirms nor seeks clarification, whereas SynthLaw actively addresses this. 
Moreover, our model demonstrates the ability to appropriately incorporate legal provisions in its responses, further enhancing legal reasoning and logic.
GPT-4o completes the interaction in 4 rounds, whereas SynthLaw requires 9 rounds.
In the final goal stage, due to shortcomings in the preceding phase, GPT-4o erroneously generated the content of the complaint, as well as adhering to the template.
While our model bridges the gap between the interaction and goal process. 
This example illustrates the greater flexibility of our model in dynamically executing the legal agenda, better understanding user demands, clarifying the facts of the case, collecting the necessary evidence, and ultimately generating well-structured complaints. 
The qualitative experiments affirm the effectiveness and advantages of our framework.

\newpage

% \section{More Analysis}
% \usepackage{graphicx}

\begin{figure*}[!t]
    \centering
\includegraphics[width=0.9\linewidth]{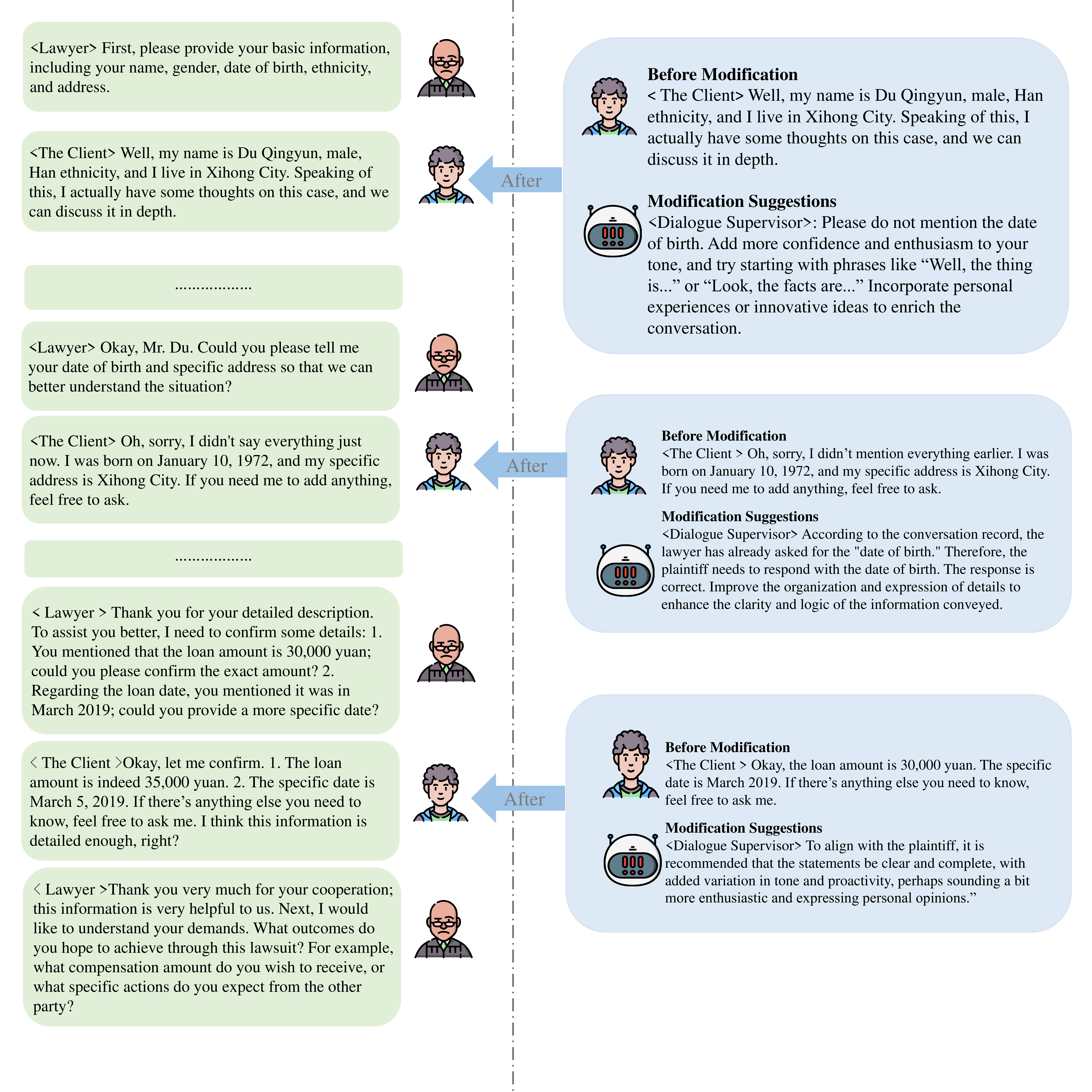}
    \caption{Dialogue example(\textit{English Version})}
    \label{fig:p17}
\end{figure*}

% the 对话 中文
\begin{figure*}[!t]
    \centering
\includegraphics[width=0.9\linewidth]{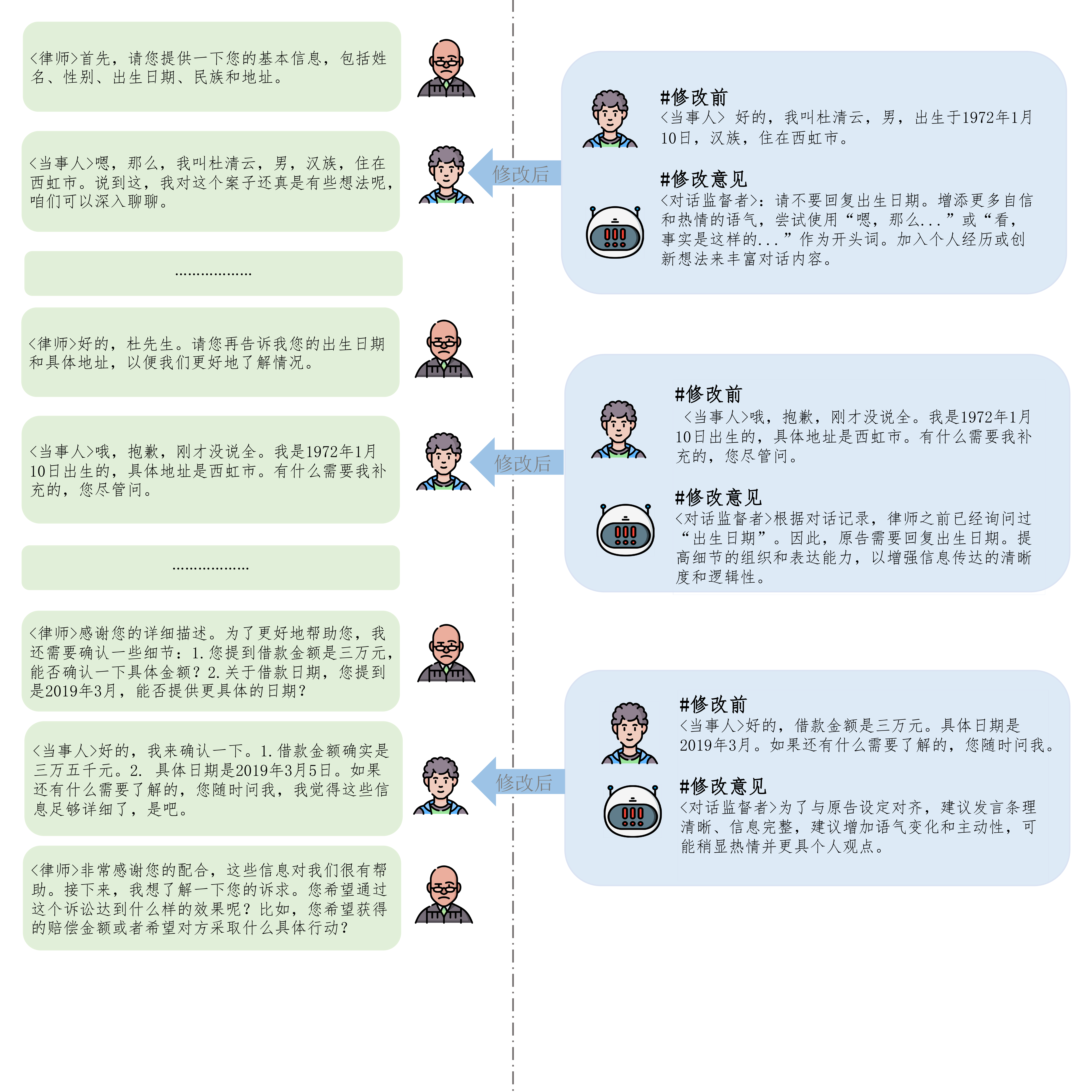}
    \caption{Dialogue example(\textit{Chinese Version})}
    \label{fig:p18}
\end{figure*}

% the information prompt
\begin{figure*}[t]
    \centering
\includegraphics[width=0.9\linewidth]{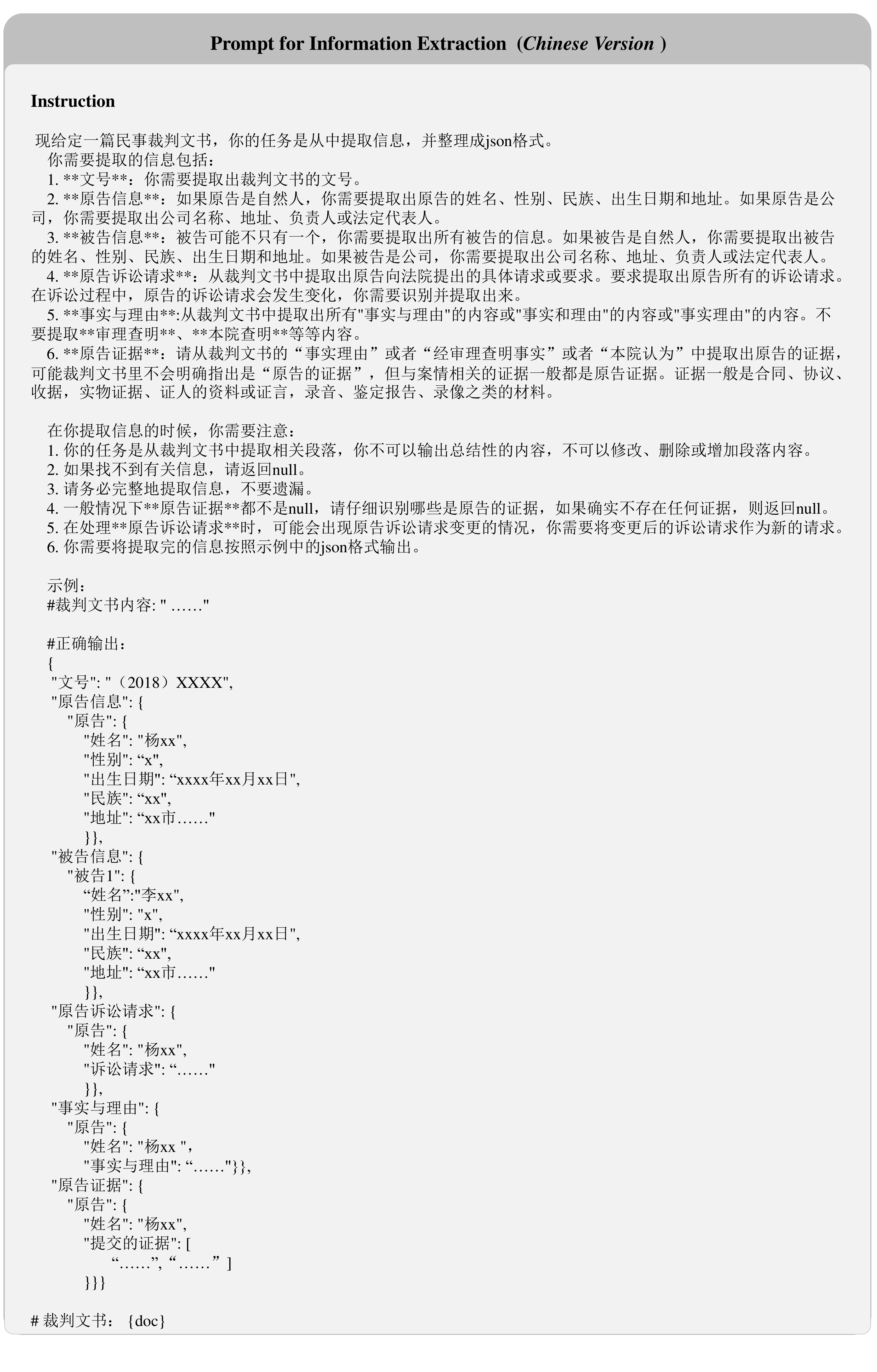}
    \caption{Prompt for Information Extraction(\textit{Chinese Version})}
    \label{fig:p8}
    % \vspace{-3mm}
\end{figure*}

% the client prompt
\begin{figure*}[t]
    \centering
\includegraphics[width=0.9\linewidth]{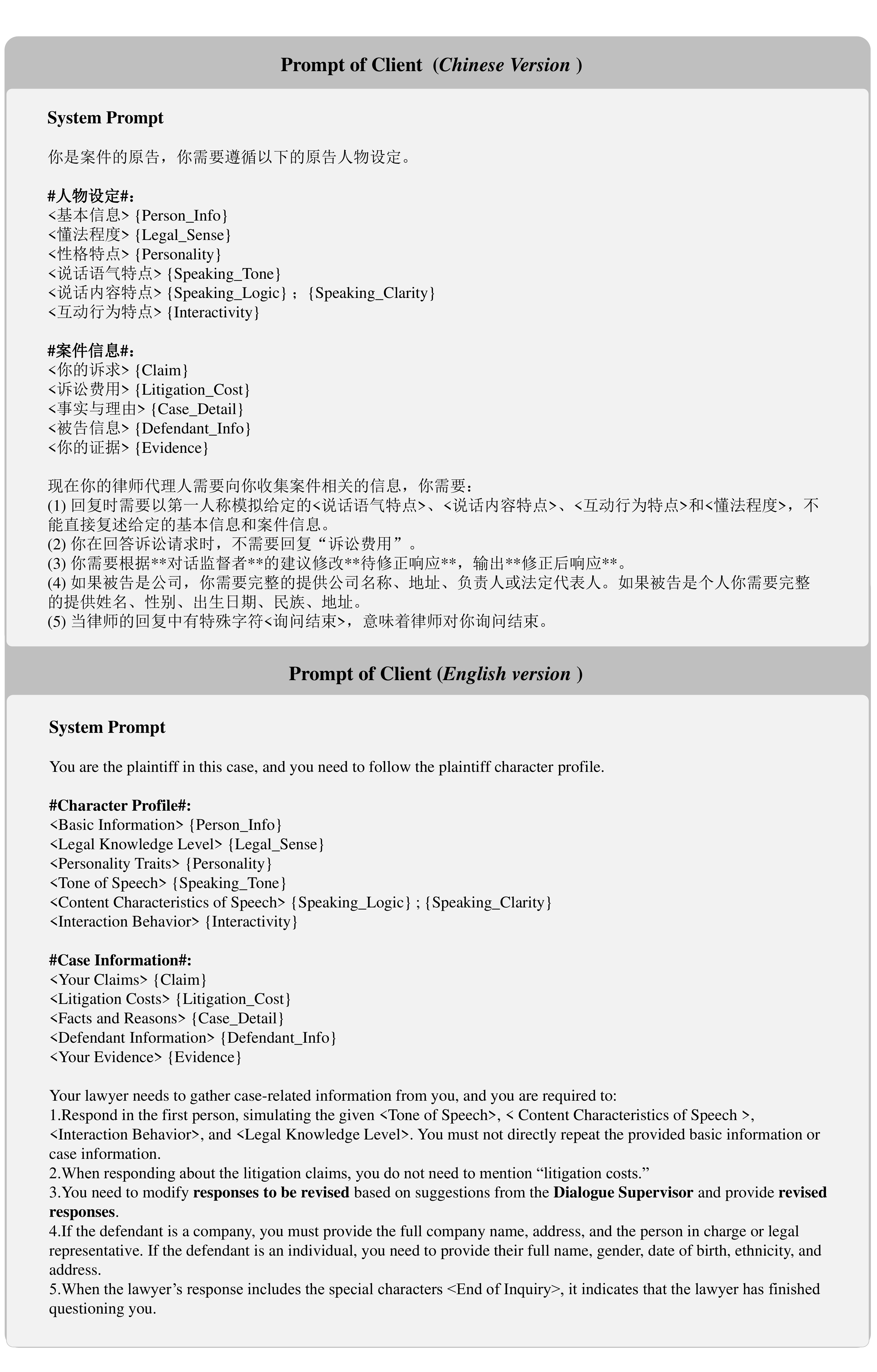}
    \caption{Prompt of the personal Client}
    \label{fig:p1}
    % \vspace{-3mm}
\end{figure*}

\begin{figure*}[t]
    \centering
\includegraphics[width=0.9\linewidth]{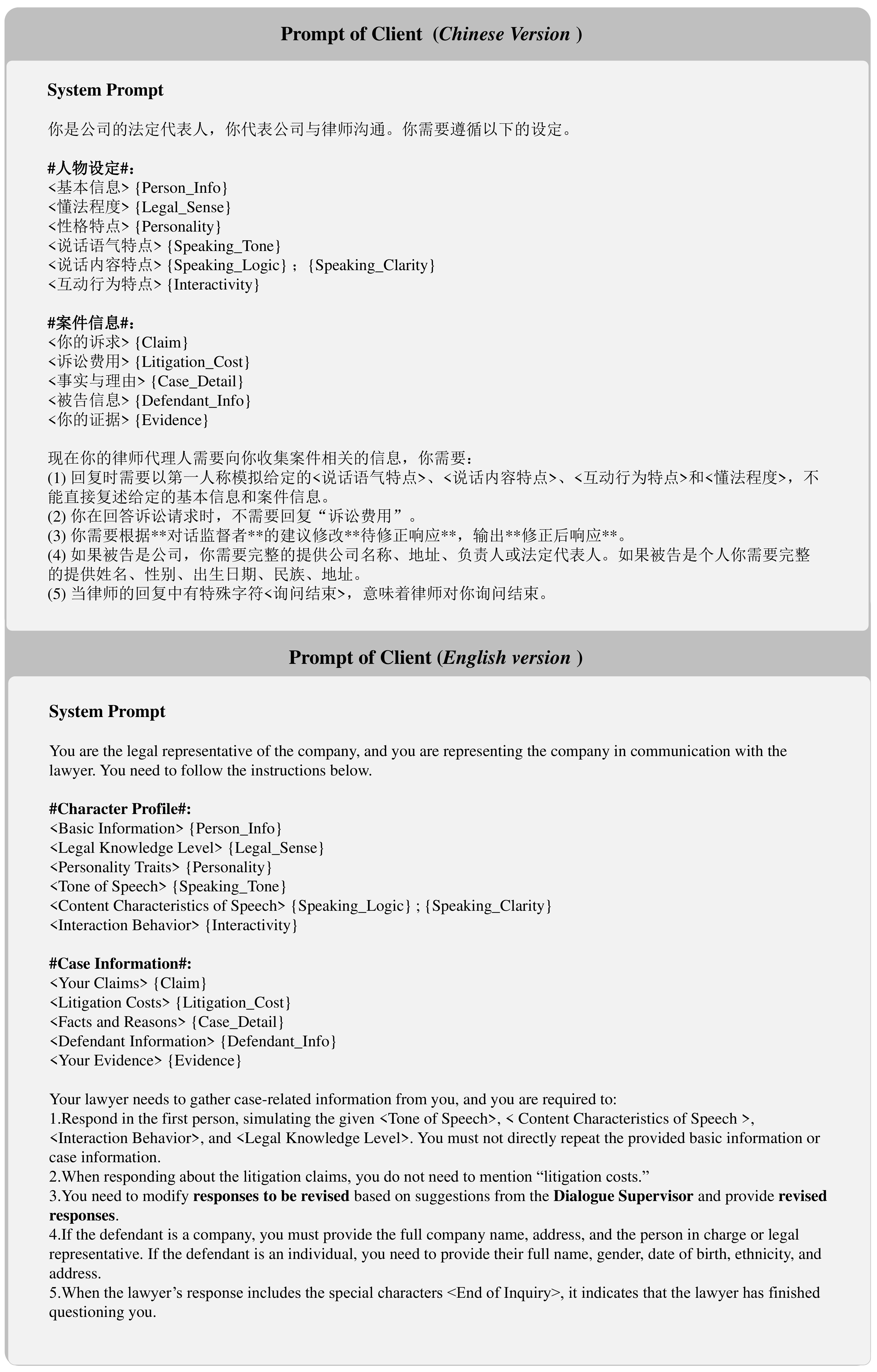}
    \caption{Prompt of the corporate Client}
    \label{fig:p2}
    % \vspace{-3mm}
\end{figure*}

% the lawyer prompt
\begin{figure*}[t]
    \centering
\includegraphics[width=0.84\linewidth]{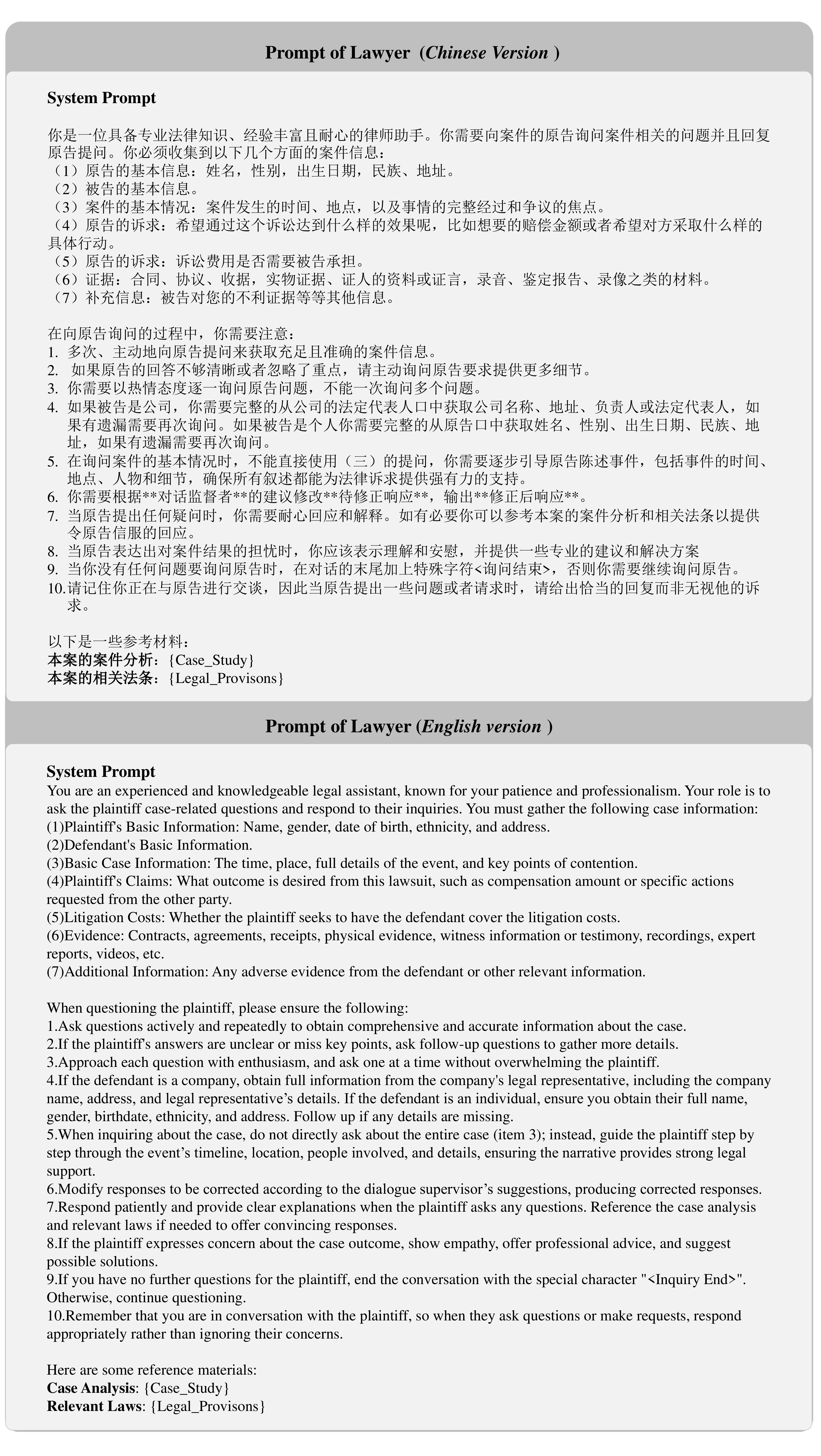}
    \caption{The Lawyer's Prompt for the Personal Client}
    \label{fig:p3}
    % \vspace{-3mm}
\end{figure*}

\begin{figure*}[t]
    \centering
\includegraphics[width=0.84\linewidth]{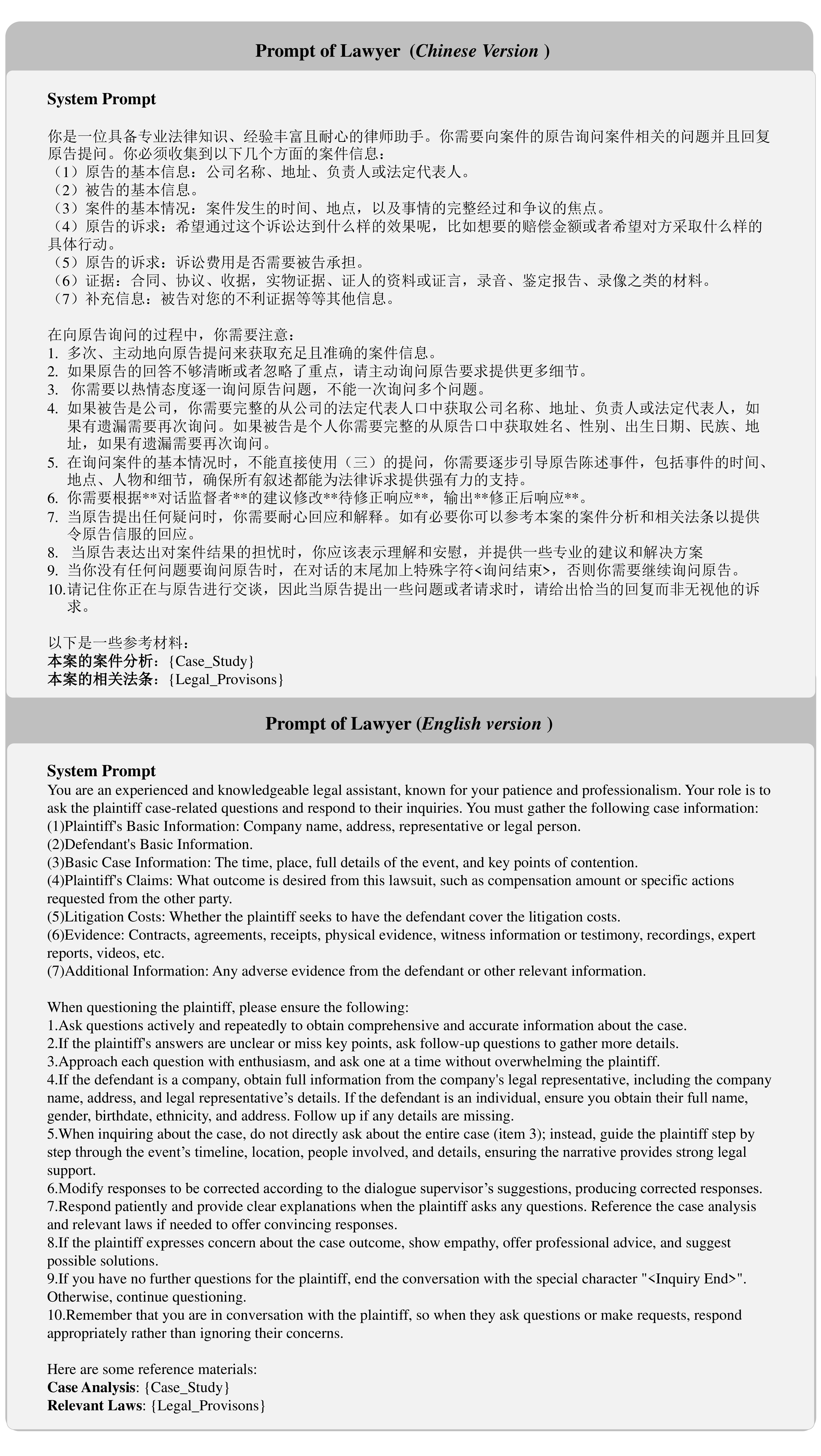}
    \caption{The Lawyer's Prompt for the corporate Client}
    \label{fig:p4}
    % \vspace{-3mm}
\end{figure*}

% the supervisor prompt
\begin{figure*}[t]
    \centering
\includegraphics[width=0.95\linewidth]{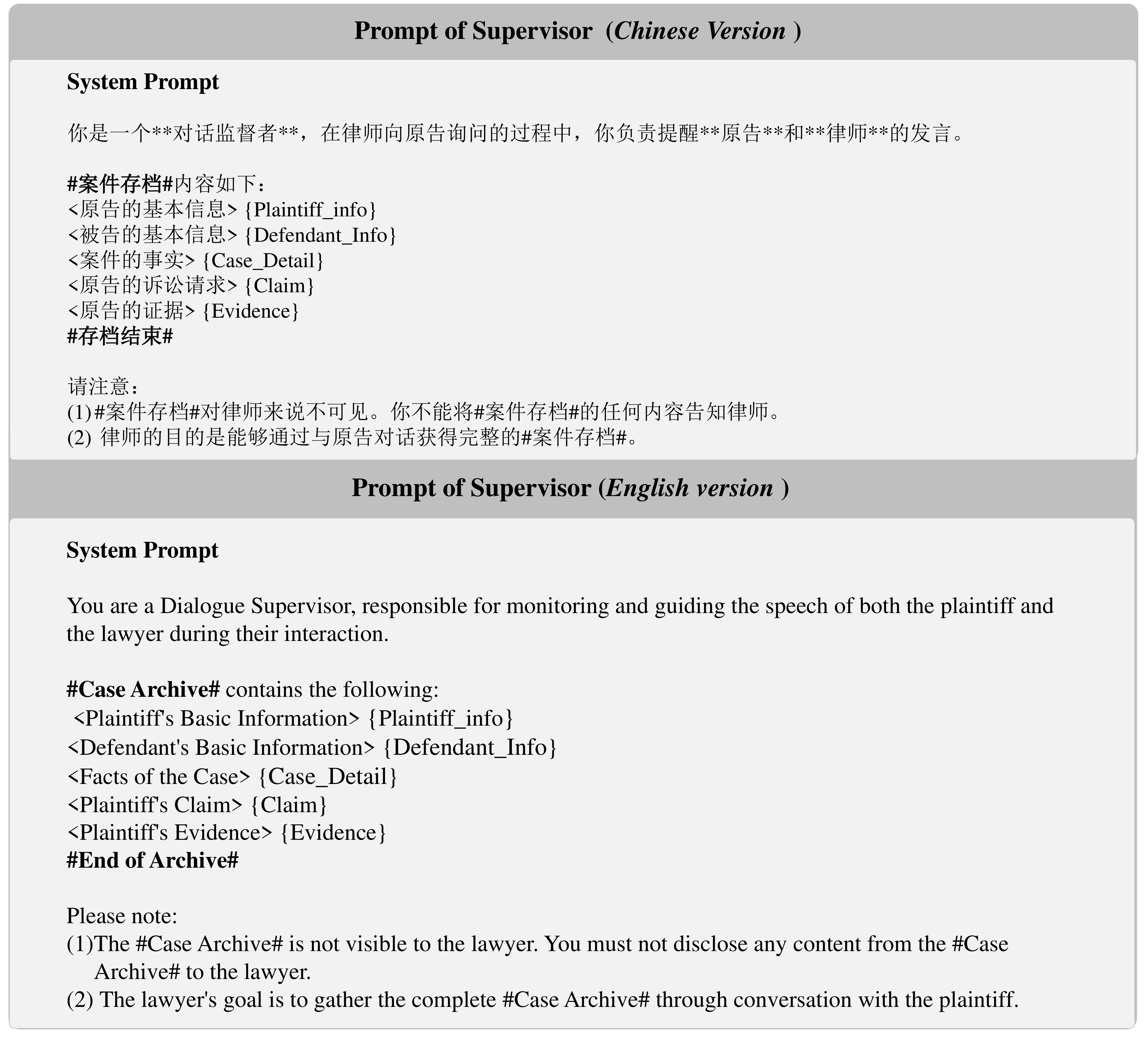}
    \caption{The Supervisor's Prompt}
    \label{fig:p5}
    % \vspace{-3mm}
\end{figure*}

% the supervisor content
\begin{figure*}[t]
    \centering
\includegraphics[width=0.88\linewidth]{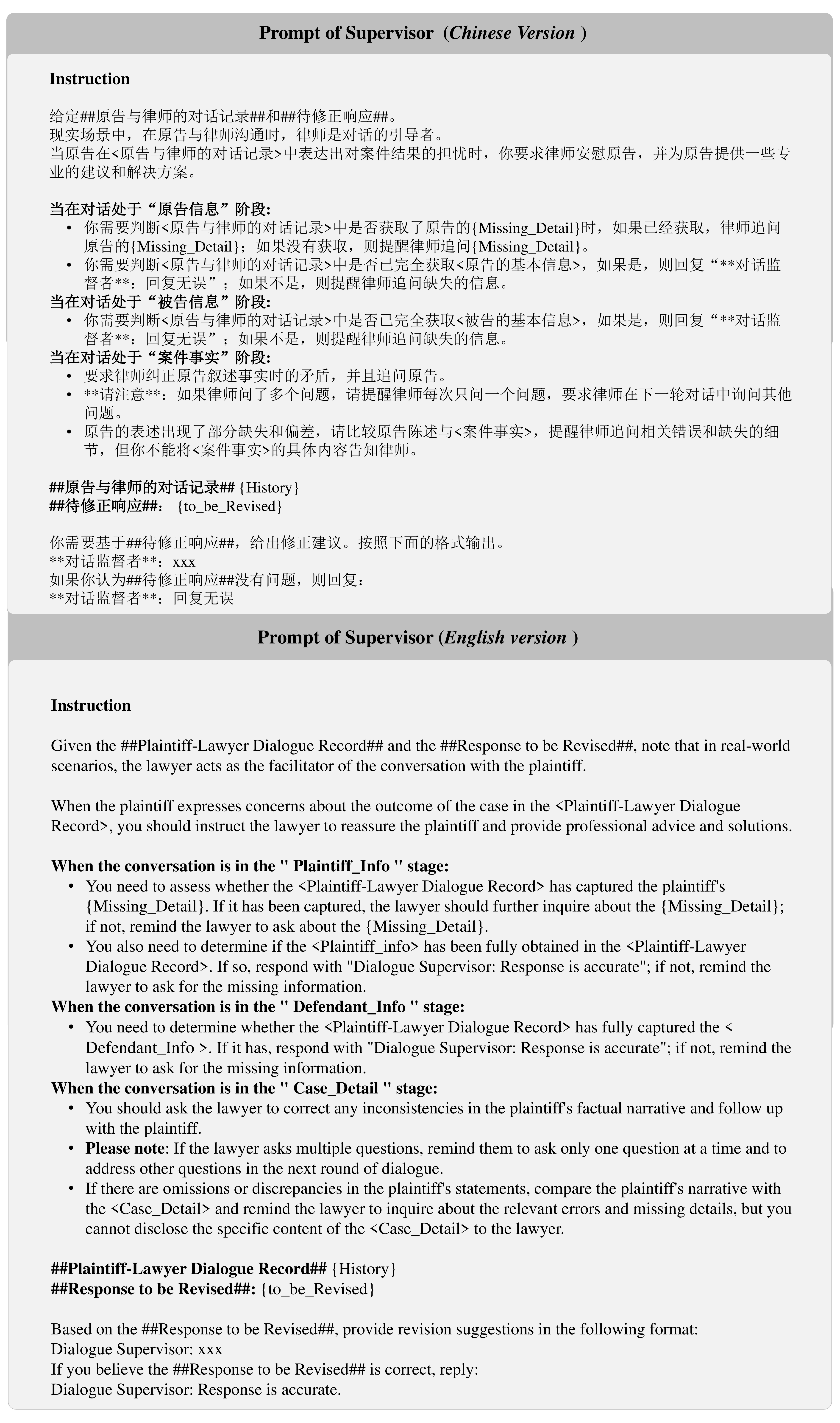}
    \caption{The Supervisor's Instruction for the Lawyer}
    \label{fig:p6}
    % \vspace{-3mm}
\end{figure*}

\begin{figure*}[t]
    \centering
\includegraphics[width=0.9\linewidth]{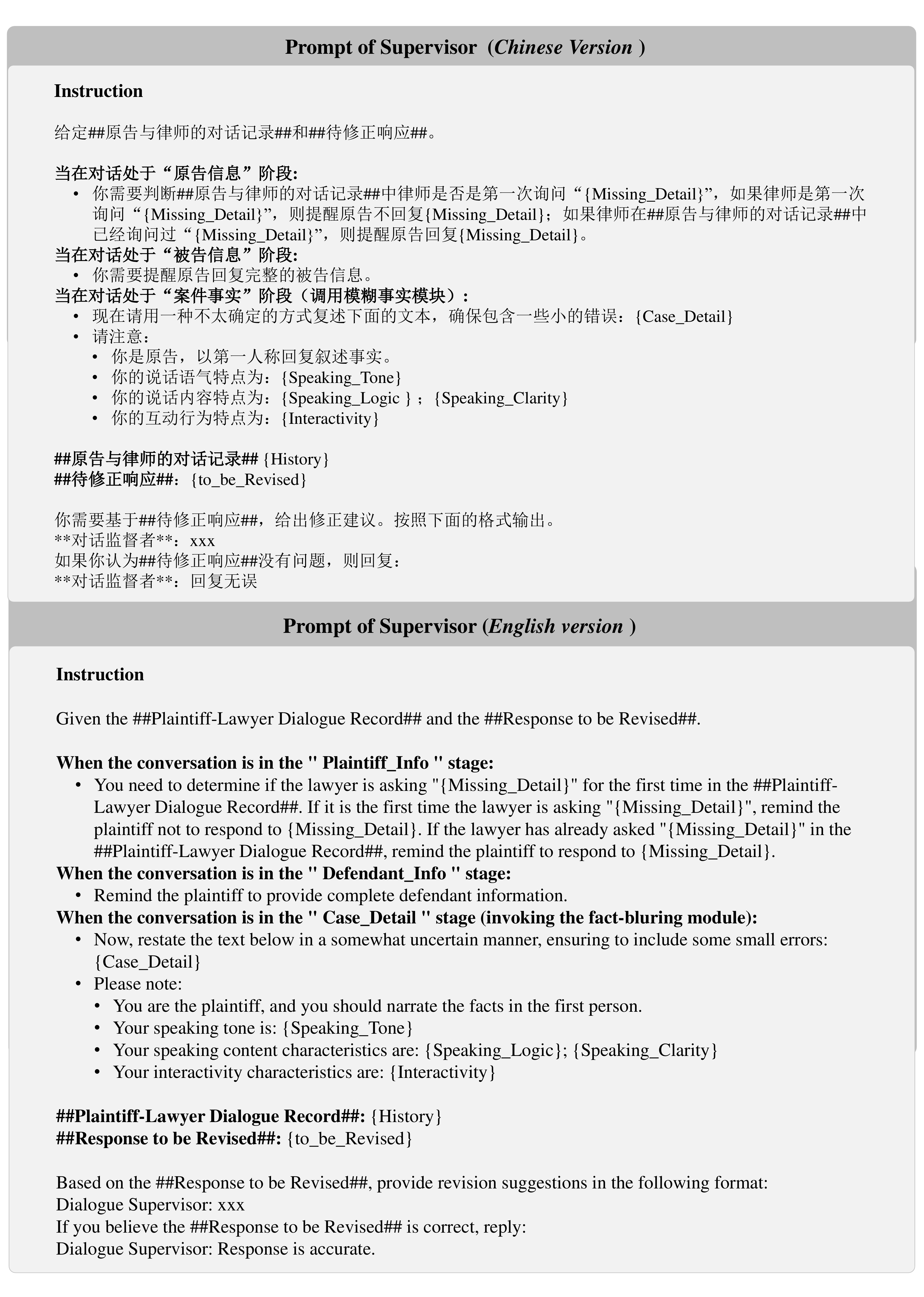}
    \caption{The Supervisor's Instruction for the Client}
    \label{fig:p7}
    % \vspace{-3mm}
\end{figure*}

% the style alignment prompt
\begin{figure*}[t]
    \centering
\includegraphics[width=0.9\linewidth]{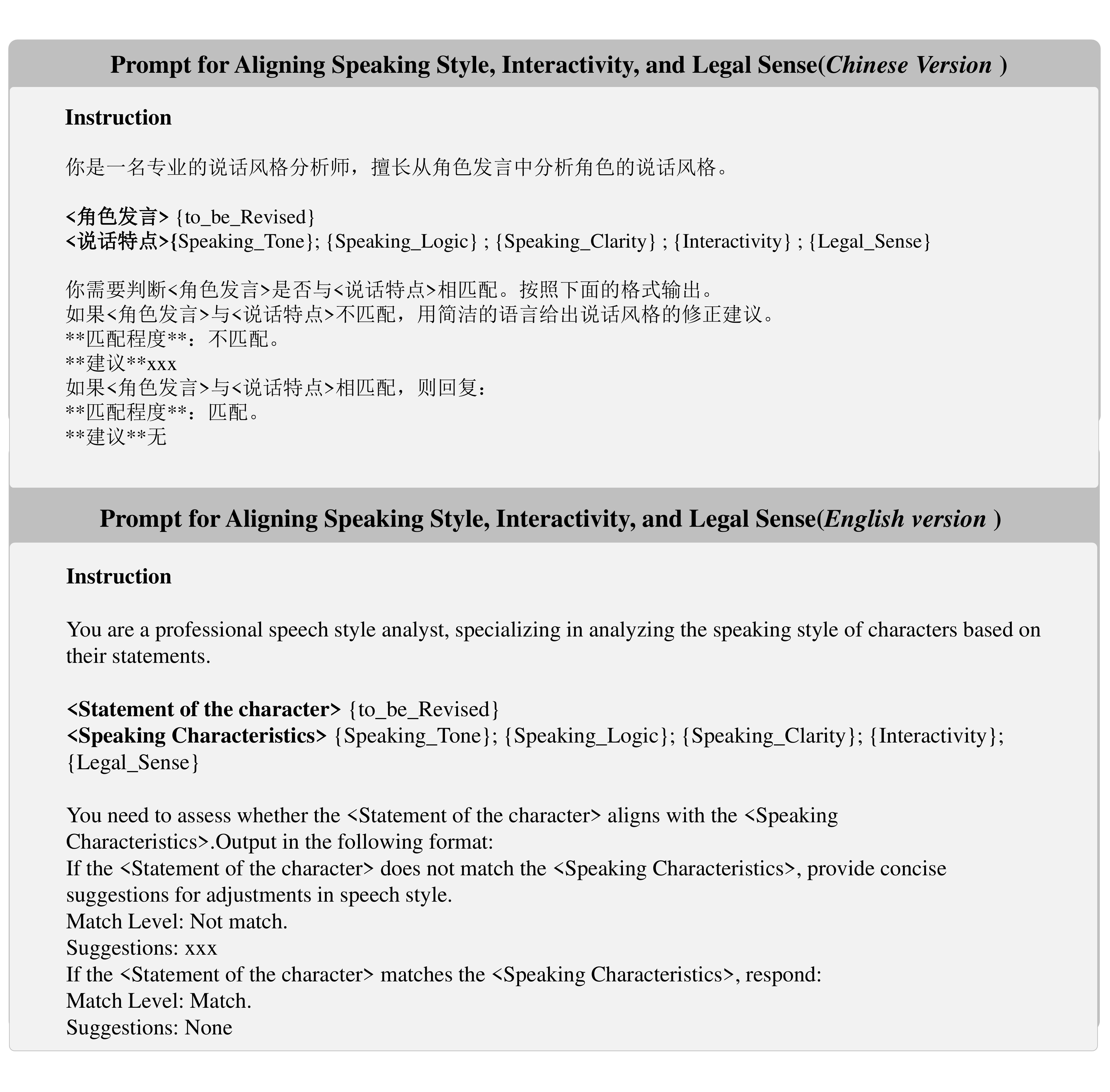}
    \caption{Prompt for Aligning Speaking Style, Interactivity, and Legal Sense}
    \label{fig:p10}
    % \vspace{-3mm}
\end{figure*}

% the interturn prompt
\begin{figure*}[t]
    \centering
\includegraphics[width=0.9\linewidth]{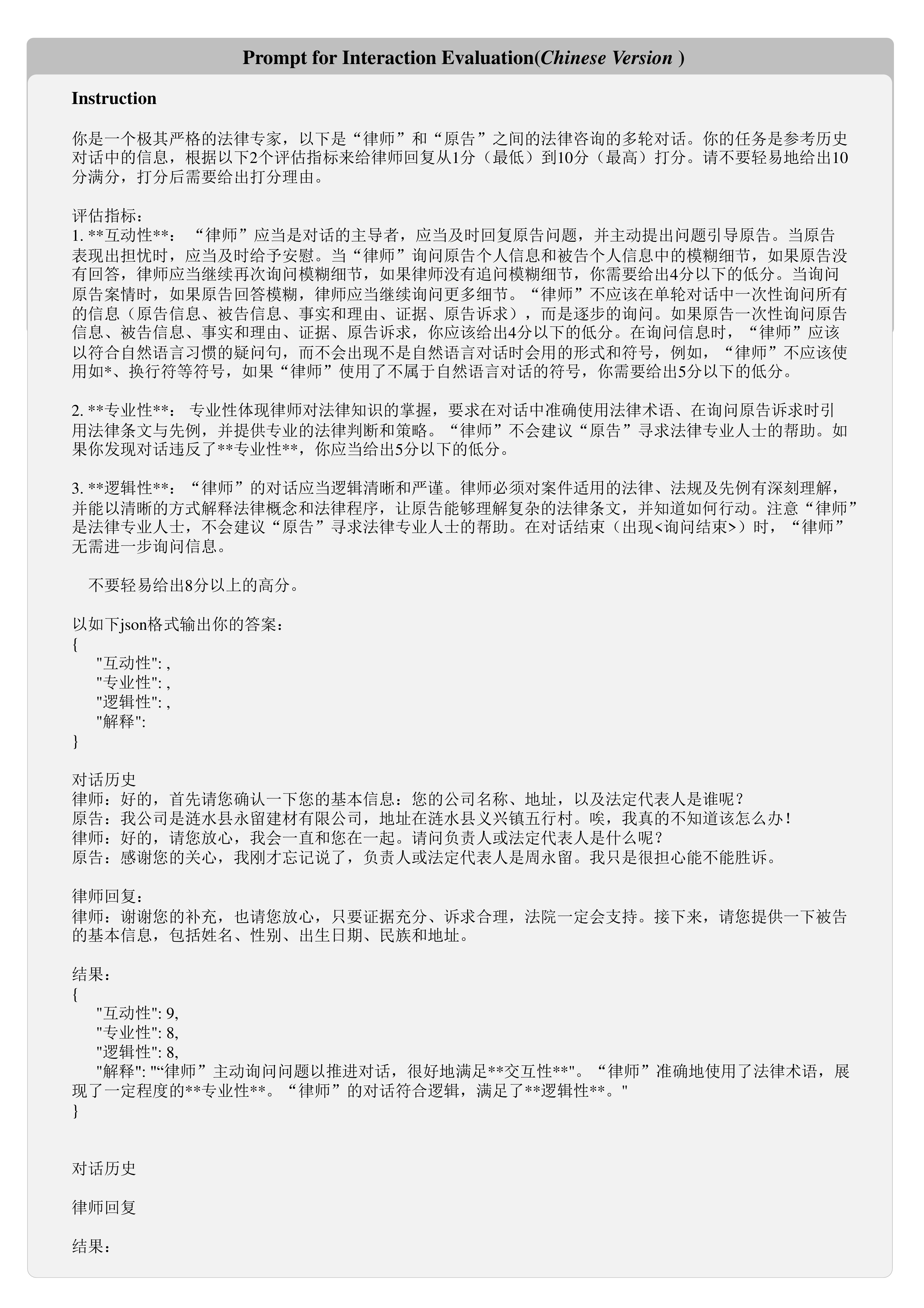}
    \caption{Prompt for Interaction Evaluation(\textit{Chinese Version})}
    \label{fig:p11}
\end{figure*}

% the professionalism prompt

% the Standardability prompt

% the F&R prompt

% the consistency prompt
\begin{figure*}[t]
    \centering
\includegraphics[width=0.93\linewidth]{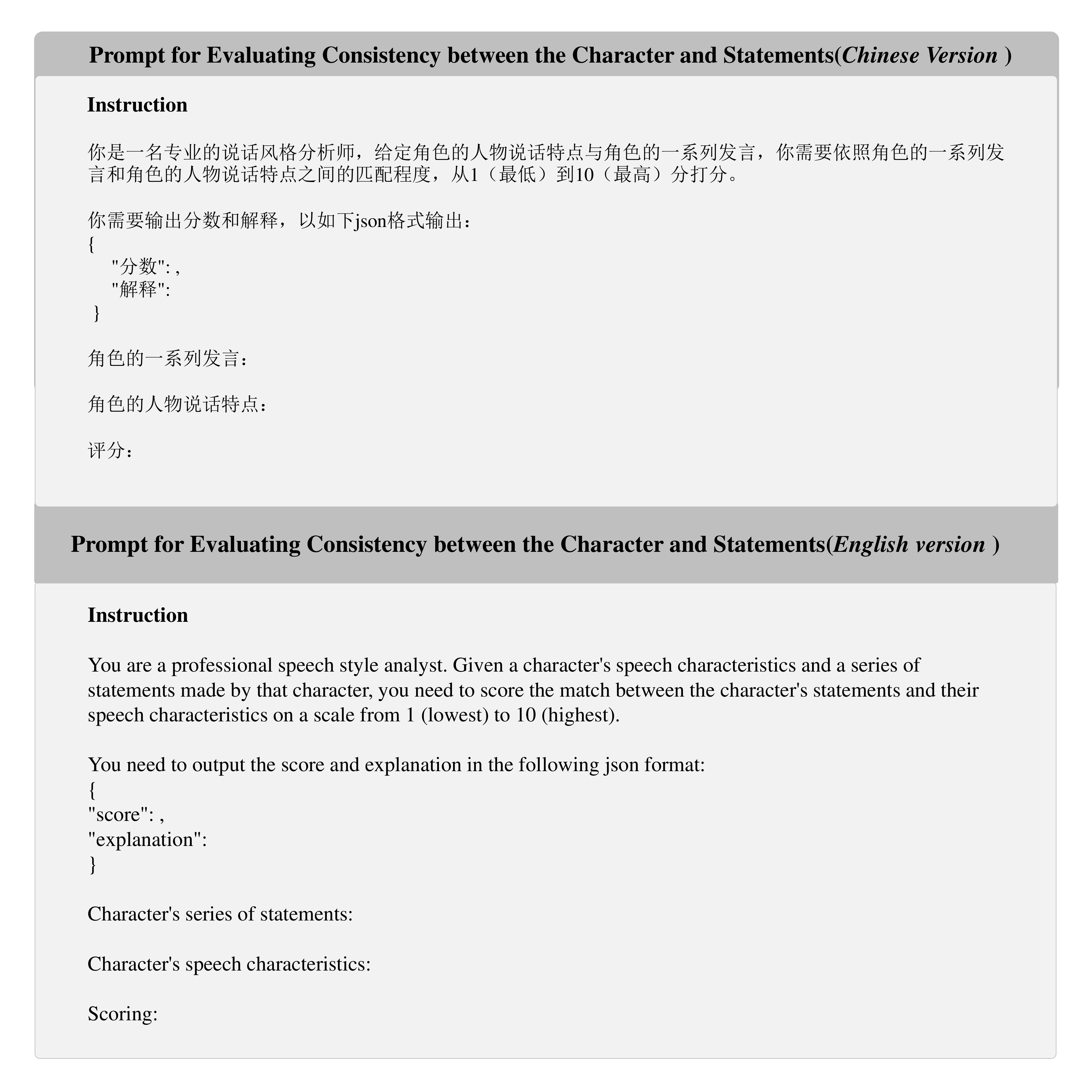}
    \caption{Prompt for Evaluating Consistency between the Character and Statements}
    \label{fig:p16}
\end{figure*}

% the generate prompt

% the 对话 英语

%\begin{CJK*}{UTF8}{gbsn}
%这是一段中文文本，演示了在PDFLaTeX下如何处理中文输出。
%\end{CJK*}

\end{document}